\documentclass{article}
\usepackage[OT1]{fontenc}
\usepackage{iclr2027_conference,times}
\iclrfinalcopy

\usepackage{amsmath,amssymb,amsfonts,amsthm}
\usepackage{mathtools}
\usepackage{booktabs}
\usepackage{graphicx}
\usepackage{xcolor}
\usepackage{hyperref}
\usepackage{url}
\usepackage{enumitem}
\usepackage{algorithm}
\usepackage{algpseudocode}
\usepackage{tikz}
\usetikzlibrary{arrows.meta,positioning,calc}

\hypersetup{
  hidelinks,
  pdftitle={Model-Aware Schedules Improve Generation via Fiberwise Optimal Transport},
  pdfauthor={Luyi Jia, Boyan Zhang, Yilun Liu, Steffen Rulands}
}

\definecolor{theoryblue}{RGB}{42,113,184}
\definecolor{theoryorange}{RGB}{213,123,18}
\definecolor{paleBlue}{RGB}{238,246,252}
\definecolor{paleOrange}{RGB}{253,244,232}
\hypersetup{colorlinks=true,citecolor=theoryblue,linkcolor=theoryblue,urlcolor=theoryblue}

\newtheorem{theorem}{Theorem}
\newtheorem{corollary}{Corollary}
\newtheorem{proposition}{Proposition}
\theoremstyle{remark}

\newcommand{\R}{\mathbb{R}}
\newcommand{\Z}{\mathbb{Z}}
\newcommand{\norm}[1]{\left\lVert #1 \right\rVert}
\newcommand{\lam}{\lambda_W}
\newcommand{\lamn}{{\lambda_W^{\mathrm{norm}}}}
\newcommand{\Nloss}{\mathcal N}
\newcommand{\kg}{\Lambda_{\mathrm g}}
\newcommand{\sig}{\boldsymbol\sigma}
\newcommand{\yy}{\mathbf y}
\newcommand{\ff}{\mathbf f}
\newcommand{\uu}{\mathbf u}
\newcommand{\btheta}{\boldsymbol{\theta}}
\newcommand{\xx}{\mathbf x}
\newcommand{\ddelta}{\boldsymbol{\delta}}

\title{A Spectral Theory of Grokking: Weight Decay induces Feature Learning}
\author{%
  Lenz Pracher$^{1,2*}$ \quad
  Pascal de Jong$^{1,*}$ \quad
  Oskar Lieshaus$^{1}$ \quad
  Alan Jeffares$^{3}$ \quad
  Steffen Rulands$^{1\dagger}$\\[3pt]
  {\normalfont $^{1}$Arnold-Sommerfeld-Center for Theoretical Physics,}\\
  {\normalfont \hphantom{$^{1}$}Ludwig-Maximilians-Universit\"at M\"unchen, Munich, Germany}\\
  {\normalfont $^{2}$Department of Applied Physics, Standford University, Standford, CA, USA}\\
  {\normalfont $^{3}$Department of Mathematics, University of Cambridge, Cambridge, UK}\\[3pt]
  {\normalfont\small
    $^{*}$Equal contribution. \qquad
    $^{\dagger}$Corresponding author (\texttt{rulands@lmu.de}).}
}

\begin{document}

\maketitle
\fancyhead{}
\renewcommand{\headrulewidth}{0pt}
\begin{abstract}
In grokking an early fit to the training data separates from a much later improvement in generalization. During this delay, training can move from a fixed neural tangent kernel (NTK) regime to one in which task-relevant kernel eigendirections continue to evolve. We provide a quantitative theory for how this transition from lazy to rich learning can produce delayed generalization. For homogeneous networks trained with squared loss and $L_2$ weight decay, we show that a finite residual remains after memorization, with larger residual fractions in target components associated with smaller NTK eigenvalues. These residuals feed back into the dynamics of the NTK itself, and projecting the resulting dynamics onto task-relevant spectral directions yields a reduced system in which residual-driven kernel growth competes with weight decay. This system predicts that the grokking timescale is controlled by the product of learning rate and weight decay, that feature learning slows logarithmically near a critical decay above which task-aligned NTK structure can no longer support generalization, and that stronger decay can prevent fitting altogether. We test these predictions in modular addition. In a homogeneous MLP, task-aligned Fourier structure continues to emerge in the NTK after training accuracy has saturated, and an 84$\times$90-grid of trained networks across varying learning rate and weight decay recovers the predicted phase geometry and inverse-product scaling of the generalization time with learning rate and weight decay. A one-block Transformer shows similar macroscopic phase structure in a 42$\times$45-grid, as well as the same transition-time scaling despite violating exact homogeneity. Together, these results provide a mechanistic derivation connecting post-fit feature learning to both the onset of generalization and its phase structure in the learning rate and weight decay plane.
\end{abstract}

\section{Introduction}
Analyzing the dynamics of generalization in deep neural networks is challenging, because optimization processes that lead to memorizing or overfitting, representation learning, and changes in validation set predictions usually occur together. However, grokking separates these timescales, as a network can reach high training accuracy, while remaining inaccurate on held-out examples for thousands of additional updates and only then generalize~\citep{power2022grokking}. This separation lets us ask what happens after the training labels are already fitted and predicted correctly. \emph{What error remains after the training labels are already fitted, how does it reshape the learned features, and what sets the delay before those changes improve held-out predictions?}

The neural tangent kernel (NTK) provides a framework for distinguishing approximately fixed-feature dynamics from feature learning. Let $\ff$ denote the network outputs, $J=\nabla_{\btheta}\ff$ the parameter Jacobian, and $K=JJ^\top$ the empirical NTK, which quantifies how network outputs adapt to changes to the parameters. For gradient flow on a differentiable data loss $\mathcal{L}_{\mathrm{data}}$, the chain rule gives
\begin{equation}
\dot{\ff}_{\rm data}=-K\nabla_{\ff}\mathcal L_{\mathrm{data}}\,,
\label{eq:intro-ntk-flow}
\end{equation}
which is architecture-independent. When $K$ remains constant, training is thus described by a fixed kernel. On the other hand, changes in $K$ reflect changes in the tangent features of the network \citep{jacot2018ntk}. The neural tangent hierarchy extends this description by expressing the evolution of $K$ in terms of higher-order tangent tensors~\citep{huang2020nth}. The residual error, the deviation between the network prediction and the true output, then enters the dynamics of $K$ itself. Hence, the neural tangent hierarchy framework links the residual that is left after saturation of the training accuracy to subsequent changes in the NTK structure that are relevant to the task at hand.

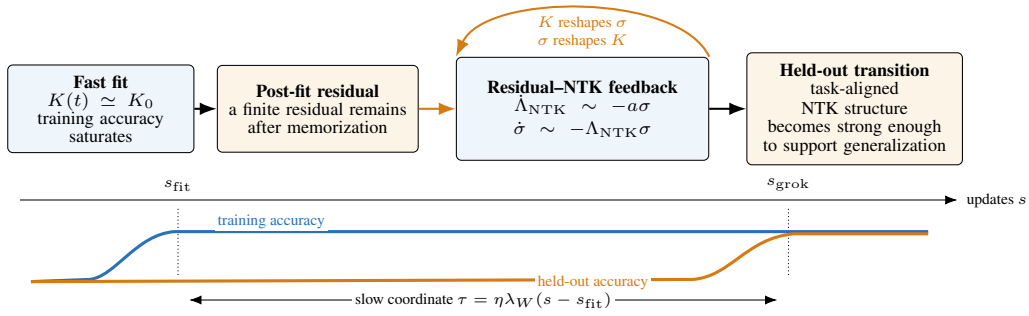
\begin{figure}[t]
\centering
\resizebox{0.98\linewidth}{!}{%
\begin{tikzpicture}[
    x=1cm,
    y=1cm,
    >=Latex,
    every node/.style={font=\scriptsize}
]
  \tikzset{
    stage/.style={
        draw,
        rounded corners=2pt,
        align=center,
        inner sep=4pt,
        minimum height=1.18cm
    },
    flow/.style={->,thick},
    feedback/.style={->,thick,theoryorange},
    note/.style={
        font=\tiny,
        fill=white,
        inner sep=1.2pt,
        align=center
    }
  }

  \node[
      stage,
      fill=paleBlue,
      text width=2.25cm
  ] (fit) at (1.20,0.62)
    {\textbf{Fast fit}\\[-1pt]
     $K(t)\simeq K_0$\\[-1pt]
     training accuracy\\[-1pt]
     saturates};

  \node[
      stage,
      fill=paleOrange,
      text width=2.45cm
  ] (res) at (4.15,0.62)
    {\textbf{Post-fit residual}\\[-1pt]
     a finite residual remains\\[-1pt]
     after memorization};

  \node[
      stage,
      fill=paleBlue,
      text width=3.15cm,
      minimum height=1.40cm
  ] (mode) at (7.75,0.62)
    {\textbf{Residual--NTK feedback}\\[-1pt]
     $\dot{\Lambda}_{\rm NTK}\sim-a\sigma$\\[1pt]
     $\dot{\sigma}\sim-\Lambda_{\rm NTK}\sigma$};

  \node[
      stage,
      fill=paleOrange,
      text width=2.65cm
  ] (margin) at (11.45,0.62)
    {\textbf{Held-out transition}\\[-1pt]
     task-aligned NTK structure\\[-1pt]
     becomes strong enough\\[-1pt]
     to support generalization};

  \draw[flow] (fit.east) -- (res.west);
  \draw[feedback] (res.east) -- (mode.west);
  \draw[flow] (mode.east) -- (margin.west);

  \draw[
      feedback,
      looseness=0.72
  ]
    (mode.north east)
    to[out=115,in=65]
    node[
        note,
        below=1.5pt
    ]
    {$K$ reshapes $\sigma$\\
     $\sigma$ reshapes $K$}
    (mode.north west);

  \draw[->,thin]
    (0.10,-0.62) --
    (12.85,-0.62)
    node[right,font=\tiny] {updates $s$};

  \draw[densely dotted]
    (2.25,-0.74) -- (2.25,-1.66);

  \draw[densely dotted]
    (10.55,-0.74) -- (10.55,-1.66);

  \node[
      font=\tiny,
      anchor=south
  ] at (2.25,-0.62)
    {$s_{\rm fit}$};

  \node[
      font=\tiny,
      anchor=south
  ] at (10.55,-0.62)
    {$s_{\rm grok}$};

  \draw[
      very thick,
      theoryblue
  ]
    (0.25,-1.73) --
    (1.05,-1.70)
    .. controls (1.40,-1.66) and (1.68,-1.09) ..
    (2.20,-1.05) --
    (12.45,-1.05);

  \draw[
      very thick,
      theoryorange
  ]
    (0.25,-1.73) --
    (9.25,-1.70)
    .. controls (9.77,-1.66) and (9.97,-1.14) ..
    (10.62,-1.08) --
    (12.45,-1.08);

  \node[
      font=\tiny,
      theoryblue,
      fill=white,
      inner sep=1pt,
      anchor=west
  ] at (2.75,-0.91)
    {training accuracy};

  \node[
      font=\tiny,
      theoryorange,
      fill=white,
      inner sep=1pt,
      anchor=west
  ] at (7.10,-1.74)
    {held-out accuracy};

  \draw[<->,thin]
    (2.40,-1.98) --
    node[
        fill=white,
        inner sep=1.2pt,
        font=\tiny
    ]
    {slow coordinate $\tau=\eta\lambda_W(s-s_{\rm fit})$}
    (10.40,-1.98);
\end{tikzpicture}
}
\caption{A fast, near-fixed-kernel stage fits the training data but leaves a finite residual under coupled weight decay. In the reduced dynamics that follow, the residual and task-aligned NTK strength evolve together. That is, the kernel reshapes the residual, and the residual reshapes the kernel, until task-aligned structure is strong enough to move held-out predictions across their decision margins. The lower panel separates the training fit from the later generalization transition and shows the slow coordinate $\tau\simeq\eta\lambda_W(s-s_{\rm fit})$.}
\label{fig:schematic}
\end{figure}

Here, we derive a post-fit mechanism for delayed generalization in homogeneous networks trained with squared loss and coupled $L_2$ weight decay. Weight decay leaves residual error after fitting, and this drives continued task-aligned evolution of the empirical NTK. For an approximately fixed NTK during the initial fit, we prove that the fraction of each target component remaining as residual error decreases monotonically with the corresponding NTK eigenvalue. The neural tangent hierarchy couples these residuals to the evolution of the NTK, allowing task-aligned tangent structure to continue developing after memorization. Modular addition provides a natural Fourier basis for projecting these dynamics onto a reduced residual-NTK system in which residual-driven growth competes with weight decay. Its adiabatic limit predicts a slow post-fit timescale controlled by the product $\eta\lambda_W$ of learning rate and weight decay, together with finite-training boundaries in the corresponding ($\eta,\lambda_W$)-phase space. Empirically, we observe the corresponding post-fit NTK organization and optimizer-space structure in a homogeneous MLP, with the same scaling of generalization time with learning rate and weight decay also appearing in a Transformer. Figure~\ref{fig:schematic} summarizes the proposed mechanism.

\paragraph{Related works.}
Delayed generalization can arise without evolving neural features. Linear estimators, Gaussian processes, and logistic models exhibit grokking under suitable conditions \citep{levi2024grokking,miller2024grokking,beck2025edge}. Most directly, ~\citet{xu2026grok} prove end-to-end grokking in overparameterized ridge regression trained by gradient descent with weight decay and derive quantitative hyperparameter dependence of the grokking time, while \citet{kim2026clock} derives an exactly solvable late-time weight-decay relaxation recovering the $(\eta\lambda_W)^{-1}$ scale in linear models. These results show that $(\eta\lambda_W)^{-1}$-type timing alone does not distinguish feature learning from fixed-feature dynamics.

Mechanistic work on neural-network grokking instead points to gradual representation change during the apparent plateau. In modular arithmetic, this includes the emergence of Fourier-structured circuits~\citep{nanda2023progress}, structured features and competition between memorizing and generalizing solutions~\citep{liu2022towards,liu2023omnigrok,varma2024explaining,merrill2023tale,ding2024survival}, and transitions from an early kernel-like regime to later feature learning \citep{kumar2024grokking,lyu2024dichotomy,mohamadi2024why,rubin2024phase,tian2026scaling}. A related line of empirical work tracks how the tangent features of the network reorganize over this transition. In particular, leading empirical-NTK eigenfunctions become increasingly task-relevant as delayed generalization emerges \citep{sanguino2024feature}, while substantial empirical-NTK movement can precede the representational changes that more closely track generalization \citep{zheng2024delays}, and empirical-NTK eigenspaces in modular-arithmetic MLPs and Transformers align increasingly with Fourier features used by the learned solution \citep{lin2025feature}. These works show that learned representations and the empirical NTK can continue to evolve during grokking, but leave open what drives this post-memorization feature learning and how the evolution of task-aligned features is linked to generalization. As pointed out by~\citet{xu2026grok}, a rigorous theoretical analysis that connects grokking to the transition from the lazy to the rich regime of training neural networks is missing.

In this work, we provide, to the best of our knowledge, the first quantitative theory linking this transition from lazy to rich training to delayed generalization. Appendix~\ref{app:related} develops the connections to other works in more detail. Specifically, we make the following contributions:
\paragraph{Contributions.}
We study delayed generalization in homogeneous networks trained with squared loss and coupled $L_2$ weight decay and derive a mechanism connecting post-fit residual error to continued task-aligned feature learning:
\begin{itemize}[leftmargin=*,itemsep=2pt,topsep=2pt]
    \item We prove that, after the initial fit, coupled weight decay leaves a finite residual in the eigendirections of an approximately fixed NTK, with the largest relative residuals in directions that are weakly represented by the current features (Theorem~\ref{thm:fast}). 
    \item We show that these residuals enter the subsequent evolution of the empirical NTK. Projecting the neural tangent hierarchy onto a task-aligned Fourier direction yields a reduced residual--NTK system in which residual-driven growth competes with weight decay (Theorem~\ref{thm:mode}).
    \item We derive a decay-controlled slow timescale for post-fit feature learning, predicting $(\eta\lambda_W)^{-1}$ scaling in the grokking regime and high-decay cutoffs where feature growth or training fit fails (Section~\ref{sec:phase}).
    \item We test these predictions in homogeneous MLPs on modular addition, observing continued Fourier organization of the NTK and the predicted weak-decay timing and optimizer-space structure. A non-homogeneous one-block Transformer shows similar macroscopic behavior (Section~\ref{sec:results}).
\end{itemize}

\section{Post-fit residuals drive tangent feature learning}\label{sec:theory}
We start by deriving the post-fit residual under coupled weight decay, and we show how it drives subsequent NTK evolution and reduce the dynamics to a task-aligned spectral mode. We then connect this mode growth to held-out generalization and the resulting optimizer-space grokking boundaries.

\subsection{Weak target modes retain larger residual fractions}\label{sec:fast}
We perform our analysis in the setting of modular arithmetic, where the task is to learn the mapping $(a,b)\mapsto c=(a+b)\bmod p$, for a given prime $p$. Inputs and targets are one-hot encoded such that the network has $p$ output coordinates. Let $\Z_p$ denote the additive group of integers modulo $p$. Its characters are $\chi_k(z)=\exp(2\pi i k z/p)$, $k\in\Z_p$, such that for output class $c$ the target can be written as
\begin{equation}
 y_c(a,b)=\frac1p\sum_{k=0}^{p-1}e^{-2\pi i k c/p}\,\chi_k(a)\chi_k(b).
\label{eq:fourier}
\end{equation}
Because the target contains only products \(\chi_k(a)\chi_k(b)\) with the same frequency \(k\), this specifies the relevant, task-aligned Fourier components that a successful model must learn to generalize. Note that the fixed-kernel residual dynamics derived below are not specific to modular addition, but hold for arbitrary datasets and targets under the homogeneous squared-loss setting. Modularity enters only when we use the task symmetry to identify an explicit Fourier basis for the subsequent spectral reduction. 

Let $n$ be the number of training examples and stack all $p$-dimensional predictions and targets into $\ff,\yy\in\R^{np}$. With residual $\sig=\ff-\yy\in\R^{np}$, the training objective can be written as
\begin{equation}
\mathcal L(\btheta)=\frac{\Nloss}{2}\norm{\sig}^2+\frac{\lam}{2}\norm{\btheta}^2\,,
\label{eq:loss}
\end{equation}
where $\Nloss$ is an arbitrary normalization factor\footnote{As shown below, the slow update time governing grokking dynamics is invariant to the specific normalization.}, and $\lam$ is the coefficient of the $L_2$ penalty. Throughout, lowercase $\lambda_W$ denotes optimizer weight decay, whereas capital $\Lambda$ denotes an NTK spectral strength or eigenvalue. We define $\lamn=\lam/\Nloss$ as the decay coefficient in normalized gradient-flow time.  One gradient-descent update with learning rate $\eta$ advances this time by $\Nloss\eta$ to first order. For $P$ trainable parameters, $J=\nabla_{\btheta}\ff\in\R^{np\times P}$ is the parameter Jacobian and $K=JJ^\top\in\R^{np\times np}$ is the empirical NTK, which measures how strongly the network output can change along any direction in output space under parameter updates. Gradient flow then leads to
\begin{equation}
\dot\btheta=-J^\top\sig-\lamn\btheta,\qquad
\dot\sig=-K\sig-\lamn J\btheta.
\label{eq:general-flow}
\end{equation}
Furthermore, if $\ff_{\btheta}$ is $D$-homogeneous under uniform parameter rescaling, Euler's identity gives $J\btheta=D\ff$. Bias-free feed-forward ReLU networks satisfy this identity, whereas biases, normalization, mixed-degree residual paths, and standard attention generally break it. Substituting this identity into Eq.~\eqref{eq:general-flow} closes the dynamics in residual space.

\begin{theorem}[Persistent ridge residual under homogeneous dynamics]
\label{thm:fast}
Let $\ff_{\btheta}$ be $D$-homogeneous and let $\lamn>0$. Gradient flow on Eq.~\eqref{eq:loss} obeys
\begin{equation}
\dot\sig=-(K+D\lamn I)\sig-D\lamn\yy.
\label{eq:hom-flow}
\end{equation}
If $K(t)\equiv K$, then
\begin{equation}
\sig(t)\longrightarrow\sig_*=-D\lamn(K+D\lamn I)^{-1}\yy.
\label{eq:ridge-fixed}
\end{equation}
Every finite-eigenvalue direction with a nonzero target projection therefore retains a nonzero residual.
\end{theorem}
For ReLU networks, the derivation holds on each time interval along the gradient-flow trajectory over which the activation pattern is fixed, and extends piecewise across activation-boundary crossings. We prove the theorem in Appendix~\ref{app:fast-proof}. Now, we diagonalize $K$, writing $K\uu_j=\Lambda_j \uu_j$, where $\uu_j$ is an eigenvector of $K$ with eigenvalue $\Lambda_j$, and denote the corresponding target and residual coefficients by $y_j=\uu_j^\top \yy$ and $\sigma_j=\uu_j^\top\sig$. The fixed-kernel limit in Eq.~\eqref{eq:ridge-fixed} then gives the equilibrium residual coefficient in the $j$th NTK eigendirection,
\begin{equation}
\sigma_{j,*}=-\frac{D\lamn}{\Lambda_j+D\lamn}\,y_j.
\label{eq:ridge-mode}
\end{equation}
This shows that for fixed $|y_j|$, the fraction of the $j$th target component that remains as residual error is larger for smaller $\Lambda_j$. These weaker modes correspond to directions that the current features represent poorly. However, in a finite network the NTK can continue to evolve after training accuracy has saturated. Therefore, we now study whether the larger relative residuals in weak task-aligned modes can drive continued task-aligned NTK evolution.

\subsection{Residuals drive task-aligned tangent growth}\label{sec:slow}
For a one-hidden-layer, $D=2$, bias-free ReLU MLP, combining the neural tangent hierarchy \citep{huang2020nth} with the two-homogeneous weight-decay terms gives
\begin{equation}
\dot\sig=-(K+2\lamn I)\sig-2\lamn\yy,\qquad
\dot K=-K^{(2)}\sig-2\lamn K.
\label{eq:nth}
\end{equation}
Here $K^{(2)}$ is a third-order tangent tensor, and $K^{(2)}\sig$ denotes contraction over its residual index,
$(K^{(2)}\sig)_{ij}:=\sum_k K^{(2)}_{ijk}\sigma_k.$
This contraction couples the residual and kernel dynamics, and again employing two-homogeneity the parameter decay term becomes $-2\lamn K$. Although this term shrinks the NTK, the residual-dependent part $-K^{(2)}\sig$ can simultaneously increase NTK strength along task-aligned directions.

To proceed, choose a normalized real Fourier direction $\uu$ with nonzero target projection and orient it so that $y=\uu^\top\yy\ge0$. We define the task-aligned NTK strength $\Lambda$ and projected residual $\sigma$ by
\begin{equation}
    \Lambda=\uu^\top K\uu,\qquad \sigma=\uu^\top\sig.
\end{equation}
$\Lambda$ measures how strongly the network can change its output along the selected task direction under parameter updates.

The Fourier structure of modular addition makes $\uu$ a natural task direction. This motivates a local one-mode approximation in which $\uu$ remains approximately an eigendirection of the NTK, $K\uu\simeq\Lambda\uu$, with weak mixing into other directions, and its dynamics are driven mainly by the residual component along $\uu$. Appendix~\ref{app:mode-closure} states these conditions explicitly. Section~\ref{sec:results} later shows the corresponding Fourier organization empirically. Projecting the hierarchy dynamics onto $\uu$ then gives
\begin{equation}
\dot{\sigma}\simeq-(\Lambda+2\lamn)\sigma-2\lamn y, \qquad
\dot\Lambda\simeq-a\sigma-2\lamn\Lambda.
\label{eq:mode}
\end{equation}
Here, $a(\btheta)$ is the local coupling between the projected residual and the residual-driven change in $\Lambda$, as defined in Appendix~\ref{app:mode-closure}. We consider $a(\btheta)>0$, qualitatively consistent with the observed post-fit emergence of task-aligned NTK structure. It implies that correcting the output along $\uu$ tends to increase the NTK strength in the same direction. This sign is a dynamical condition rather than a consequence of homogeneity or symmetry. Equation~\eqref{eq:mode} then describes a feedback. For $\sigma<0$, larger $\Lambda$ accelerates residual relaxation, while the remaining residual drives $\Lambda$ upward.

We further assume that $a(\btheta)$ is locally constant over the post-fit interval considered. Appendix~\ref{app:mode-closure} shows that once the residual is close to its instantaneous ridge value, weight decay does not introduce an additional fast timescale for $a$, supporting this approximation in the slow regime considered below. After the fast ridge relaxation, $\sigma<0$ for $y>0$, so $-a\sigma>0$ and the residual increases the NTK strength along the selected task direction. Under the locally constant-$a$ approximation, the fixed points of Eq.~\eqref{eq:mode} and their stability can be characterized exactly.

\begin{theorem}[Spectral selection in the reduced tangent-mode system]
\label{thm:mode}
Under the locally constant-$a$ approximation, for $\lamn>0$, $a>0$, and $y\ge0$, Eq.~\eqref{eq:mode} has one equilibrium with $\Lambda\ge0$:
\begin{equation}
\Lambda_*=-\lamn+\sqrt{\lamn^2+ay},\qquad
\sigma_*=-\frac{2\lamn}{a}\Lambda_*.
\label{eq:fixed}
\end{equation}
The equilibrium is locally asymptotically stable. It satisfies $\Lambda_*=0$ for $y=0$ and $\Lambda_*>0$ for $y>0$.
\end{theorem}
We give the fixed-point and stability calculations in Appendix~\ref{app:mode-proof}. Within the reduced system, unsupported directions decay, while positive residual-to-NTK coupling sustains NTK strength for target-aligned directions. Under the adiabatic approximation that residual relaxation is faster than mode motion, we set $\dot\sigma\simeq0$ at the current $\Lambda$ and obtain
\begin{equation}
\sigma_{\rm ad}(\Lambda)=-\frac{2\lamn y}{\Lambda+2\lamn},\qquad
\dot\Lambda=2\lamn\left(\frac{ay}{\Lambda+2\lamn}-\Lambda\right).
\label{eq:slow}
\end{equation}
This shows that mode evolution occurs on a normalized-time scale $1/\lamn$. Since $t\simeq\Nloss\eta s$ after $s$ updates, we define the normalization-invariant slow time
\begin{equation}
\tau:=\lamn t\simeq\eta\lam s\,.
\label{eq:clock}
\end{equation}
Here, $s$ is the optimizer update index and $\eta\lambda_W$ is the per-update decay scale. If $s_{\rm fit}$ denotes the end of the initial fitting stage, the subsequent slow-time increment is $\Delta\tau\simeq\eta\lambda_W(s-s_{\rm fit})$. Thus the reduced dynamics predict the leading scaling of post-fit transition times with $(\eta\lambda_W)^{-1}$.

\subsection{Grokking boundaries in the $(\eta,\lambda_W)$ plane}\label{sec:phase}
The same adiabatic calculation shows how the task-aligned NTK strength $\Lambda$ controls how much of the corresponding target component is expressed in the network output. Since $\sigma=\uu^\top(\ff-\yy)$, the projected output coefficient $f=\uu^\top\ff$ satisfies
\begin{equation}
f=y+\sigma_{\rm ad}(\Lambda)=g(\Lambda)y,\quad g(\Lambda)=\frac{\Lambda}{\Lambda+2\lamn}.
\label{eq:ridge-gain}
\end{equation}
Hence, increasing $\Lambda$ along a task-aligned mode increases the corresponding component of the training outputs continuously. This training coordinate is related to generalization through the evolution of held-out margins, the difference between the correct-class logit and the largest competing logit, along the same trajectory. Because modular addition is represented by several matched Fourier components as shown in Eq.~\eqref{eq:fourier}, these margins can change smoothly as the corresponding components grow, while the predicted label changes only when a margin crosses zero. If many held-out examples cross that threshold at similar mode strengths, smooth spectral growth can therefore produce a sharp rise in validation accuracy. Appendix~\ref{app:boundary-margins} formalizes this connection using the empirical distribution of the mode strengths at which individual held-out examples change class.

The scalar dynamics in Eq.~\eqref{eq:mode} track a single task-aligned Fourier mode and do not resolve the individual held-out margins that produce the sharp accuracy transition. We therefore now replace the collection of margin crossings by a threshold $\kg$ on the task-aligned NTK strength $\Lambda$. Reaching $\kg$ represents reaching a chosen held-out-accuracy threshold. We denote the NTK strength at the beginning of the slow stage by $\Lambda_0$. Further, we consider a total of $T$ training updates, and recall the stable equilibrium $\Lambda_*$ from Eq.~\eqref{eq:fixed}. Since Eq.~\eqref{eq:slow} gives $\dot\Lambda>0$ for $\Lambda<\Lambda_*$, a trajectory with $\Lambda_0<\kg<\Lambda_*$ crosses the threshold $\kg$ in finite time. Separating variables in Eq.~\eqref{eq:slow} and integrating then gives the boundary in the $(\eta,\lam)$ plane between trajectories that reach $\kg$ within $T$ updates and those that remain below it.

\begin{corollary}[Finite-training grokking boundary]
\label{cor:boundary}
Under the one-mode, adiabatic, and gradient-flow approximations,
\begin{equation}
\eta_*(\lam)=\frac{1}{2\lam T}
\int_{\Lambda_0}^{\kg}
\frac{\Lambda+2\lamn}{ay-\Lambda^2-2\lamn\Lambda}\,d\Lambda\,.
\label{eq:boundary}
\end{equation}
At small normalized decay and away from the fixed-point boundary, the leading dependence is $\eta_*\propto1/\lam$.
\end{corollary}
A derivation and closed form solution of this integral can be found in Appendices~\ref{app:boundary-slow}--~\ref{app:boundary-closed}. Equation~\eqref{eq:boundary} predicts that increasing $\Lambda_0$ removes a positive portion of the crossing-time integral and thereby shortens the delay before grokking occurs. We elaborate on this prediction and provide an empirical test in Appendices~\ref{app:alignment-prediction}--~\ref{app:alignment-experiment}. This finite-training grokking boundary is obtained under the assumption that the threshold is reachable in the first place, $\kg<\Lambda_*$. Whether or not this is possible follows from the limiting case $\kg=\Lambda_*$, which gives
\begin{equation}
\lam^{\mathrm{mode}}=\Nloss\,\frac{ay-\kg^2}{2\kg},
\label{eq:featurecut}
\end{equation}
provided $ay>\kg^2$. For $\lam<\lam^{\mathrm{mode}}$, the equilibrium lies above $\kg$ so the mode can reach $\kg$ in finite time, with Corollary~\ref{cor:boundary} determining whether this occurs within $T$ updates. For $\lam\geq\lam^{\mathrm{mode}}$ the threshold cannot be crossed in finite time. As $\lam$ approaches this cutoff from below, the net growth rate of the task-aligned mode at $\Lambda=\kg$ approaches zero, causing $\eta_*(\lam)$ to diverge logarithmically as $\lam\to\lam^{\mathrm{mode}}$ (Appendix~\ref{app:boundary-closed}). This shows that as $\lam$ approaches $\lam^{\mathrm{mode}}$ from below, reaching $\kg$ and thereby generalizing within a fixed update budget requires an increasingly large learning rate. In the $(\eta,\lam)$ plane, this produces an upward turn of the finite-time boundary near the cutoff at $\lam^{\mathrm{mode}}$.

A further constraint arises from fitting the training data. This boundary depends on the relative gains of multiple output modes and is therefore not fixed exactly by the one-mode dynamics. To retain a scalar description, we summarize the onset of fitting failure by the gain of an effective training mode. Requiring $g(\Lambda_{\mathrm{fit}})\ge g_{\min}$ gives
\begin{equation}
\lam\le\Nloss\,\frac{\Lambda_{\mathrm{fit}}(1-g_{\min})}{2g_{\min}},
\label{eq:fitcut}
\end{equation}
This provides an approximately vertical effective decay scale beyond which the network no longer reaches the chosen training-accuracy criterion.

Finally, sufficiently large learning rates encounter the discrete-time stability edge of the approximately fixed-kernel dynamics. As derived in Appendix~\ref{app:stability}, linear stability requires
\begin{equation}
\eta<\frac{2}{\Nloss\Lambda_{\max}(K)+2\lam}\,,
\label{eq:stabilitycut}
\end{equation}
where $\Lambda_{\max}(K)$ is the largest eigenvalue of the frozen NTK. This is the frozen-kernel analogue of the usual edge-of-stability condition~\citep{cohen2021gradient}. Because the NTK subsequently evolves, it provides only a local estimate of the large-$\eta$ boundary of the full training dynamics. Appendices~\ref{app:boundary-closed}--\ref{app:stability} give the derivations and validity conditions for the fitting and stability bounds.

To summarize, these constraints predict the following organization of the $(\eta,\lam)$ plane. When the training data are fitted and $\kg$ is reachable but the finite-training condition in Corollary~\ref{cor:boundary} is not satisfied, the network remains in the memorization regime over the available training interval. Grokking is possible when the training fit succeeds (Eq.~\eqref{eq:fitcut}), the task-aligned NTK strength can reach $\kg$ (Eq.~\eqref{eq:featurecut}), the threshold is crossed within $T$ updates (Eq.~\eqref{eq:boundary}), and gradient descent remains stable (Eq.~\eqref{eq:stabilitycut}). Increasing weight decay can eventually prevent either task-mode growth or training fit, while sufficiently large learning rates produce a separate instability boundary.

\section{Experiments}\label{sec:results}
We now test these predictions and measure post-fit NTK reorganization as well as the $(\eta,\lambda_W)$-plane boundaries, and whether similar transition-time behavior appears in Transformers. Appendix~\ref{app:protocols} provides experimental details, and all results can be reproduced from our GitHub page\footnote{The code will be made publicly available upon publication.}.

\subsection{Task-aligned NTK structure develops after fitting}\label{sec:evidence}
\begin{figure}[t]
\centering
\includegraphics[width=0.96\linewidth]{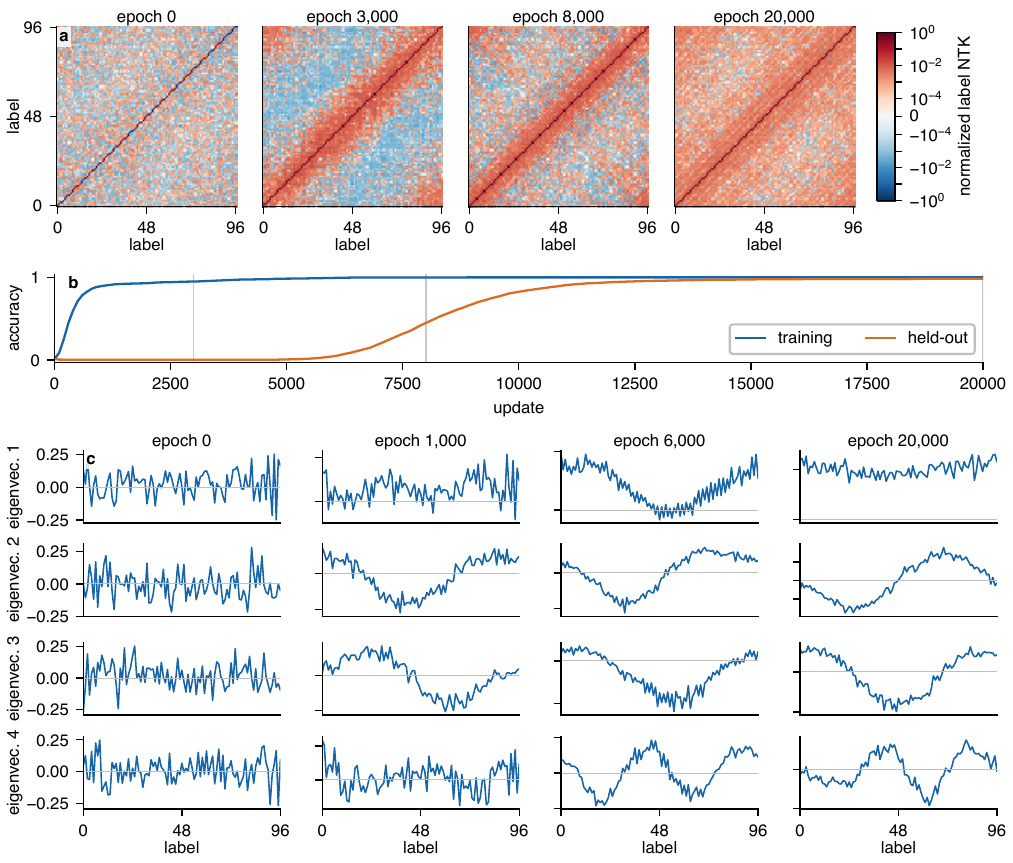}
\caption{Post-fit change in the normalized label-space NTK for a bias-free MLP on addition modulo $97$. \textbf{(a)} We project the sample NTK onto the true-label, average by output label, and normalize the resulting $97\times97$ label kernel. \textbf{(b)} Training and held-out accuracy for the same run. \textbf{(c)} Four leading eigenvectors of the same normalized label kernel, showing Fourier-like standing waves that sharpen after training accuracy is already high.}
\label{fig:ntk_accuracy_eigenvectors}
\end{figure}
We trained a bias-free one-hidden-layer ReLU MLP on addition modulo $97$, using half of the samples for training and the remainder for evaluation. We computed the empirical NTK, projected it onto the true-label output of each sample, and averaged the resulting entries by output label. This isolates tangent feature organization with respect to the task labels rather than individual examples. To further isolate structure from changes in overall NTK scale, we normalized the resulting $97\times97$ kernel. Figure~\ref{fig:ntk_accuracy_eigenvectors}a shows that periodic structure in this kernel strengthens between updates $3{,}000$ and $20{,}000$, while Figure~\ref{fig:ntk_accuracy_eigenvectors}b shows that training accuracy has already saturated during much of this evolution. In Figure~\ref{fig:ntk_accuracy_eigenvectors}c, the leading eigenvectors dynamics show the emergence of standing-wave profiles in label space, consistent with the Fourier task structure in Eq.~\eqref{eq:fourier}. Their continued organization after fitting is also qualitatively consistent with the post-fit growth of task-aligned NTK modes described by Eq.~\eqref{eq:mode}.

\subsection{Grokking boundaries in the MLP optimizer plane}\label{sec:phase-results}
\begin{figure}[t]
\centering
\includegraphics[width=\linewidth]{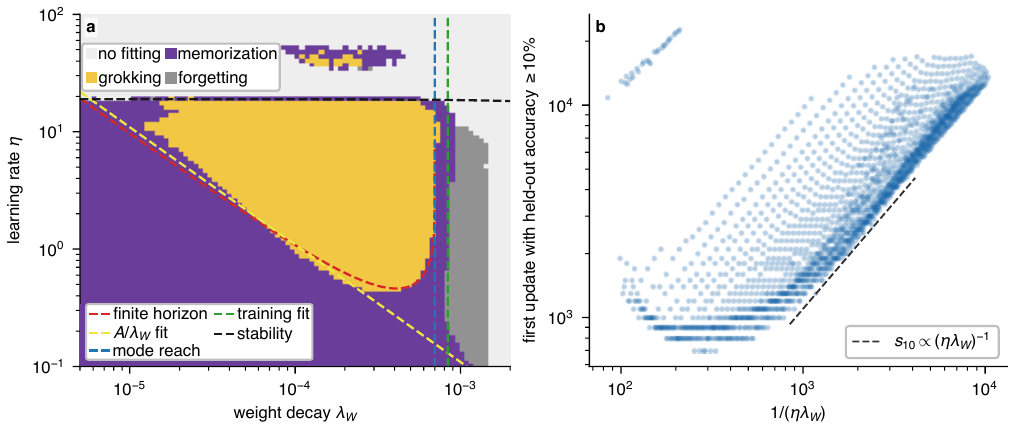}
\caption{Optimizer-plane structure for the homogeneous MLP on addition modulo $23$. \textbf{(a)} Outcomes after $24{,}000$ updates on an $84\times90$ learning-rate/decay grid. The overlays show the calibrated finite-time boundary, a separately fitted inverse-decay relation, the mode-reachability cutoff, the training-fit cutoff, and the calibrated frozen-kernel stability form. \textbf{(b)} First held-out $10\%$ crossing among runs that eventually meet the $90\%$ grokking criterion, plotted against $(\eta\lambda_W)^{-1}$. The dashed segment indicates the predicted scaling $s_{10}\propto(\eta\lambda_W)^{-1}$ and is not fitted to the data.}
\label{fig:mlp_phase_and_clock}
\end{figure}
Next, we trained an (84$\times$90)-grid of MLPs for varying $(\eta,\lambda_W)$ on addition modulo $23$ for a fixed total of $24{,}000$ updates. Figure~\ref{fig:mlp_phase_and_clock} shows the resulting $(\eta$,$\lambda_W)$-plane, categorized into memorizing, grokking, forgetting, and non-fitting runs, as described in Appendix~\ref{app:protocol-mlp}. At small $\eta\lambda_W$, the task-aligned NTK strength $\Lambda$ does not have enough time to reach the generalization threshold $\kg$ within the fixed training budget, producing the broad memorization region in Figure~\ref{fig:mlp_phase_and_clock}a. The red curve fits the finite-training grokking boundary from Eq.~\eqref{eq:boundary}. Note that since the fit-parameters are inferred from the boundary, they are not direct measurements of $\Lambda(t)$. Hence, this comparison tests the predicted optimizer dependence of the crossing time rather than the absolute amount of task-aligned NTK growth. At weak decay, it follows the transition from memorization to grokking and closely tracks the yellow $\eta\propto\lambda_W^{-1}$ fit. As $\lambda_W$ increases, the finite-time curve turns upward toward the blue mode-reachability cutoff from Eq.~\eqref{eq:featurecut}, beyond which the task-aligned NTK strength cannot reach the required threshold. The green line shows the training-fit cutoff from Eq.~\eqref{eq:fitcut}, while the black line corresponds to the frozen-kernel stability condition in Eq.~\eqref{eq:stabilitycut}, fitted to the observed large-learning-rate boundary. The full fitting procedures for all curves are given in Appendix~\ref{app:protocol-mlp-calibration}.

Figure~\ref{fig:mlp_phase_and_clock}b tests the predicted slow-time scaling. For every run that reaches $90\%$ held-out accuracy, we plot the first update $s_{10}$ at which that accuracy reaches $10\%$ against $(\eta\lambda_W)^{-1}$. Equation~\eqref{eq:clock} predicts $s_{10}\propto(\eta\lambda_W)^{-1}$ if the transition is governed by the slow time $\tau\simeq\eta\lambda_W s$. Indeed, the observed crossing times follow this scaling across the grokking region.

\subsection{Grokking boundaries and transition-time scaling in a Transformer}\label{sec:transformer}
\begin{figure}[t]
\centering
\includegraphics[width=\linewidth]{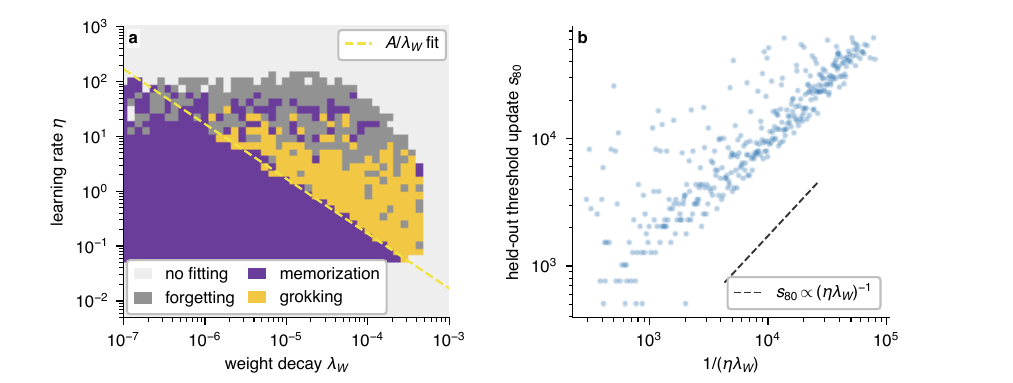}
\caption{Phase and transition-time measurements for a one-block Transformer on the same modulo-$23$ split. \textbf{(a)} Outcomes on a $42\times45$ optimizer grid with an inverse-decay fit to the main memorization--grokking boundary. \textbf{(b)} First held-out $80\%$ crossing among runs that ultimately grok at the $90\%$ criterion, plotted against $(\eta\lambda_W)^{-1}$. The dashed segment indicates the scaling $s_{80}\propto(\eta\lambda_W)^{-1}$.}
\label{fig:transformer_phase_and_clock}
\end{figure}
Biases, LayerNorm, residual paths, and attention all break the global homogeneity used in Theorems~\ref{thm:fast} and \ref{thm:mode}. However, differentiating $K=JJ^\top$ under gradient flow still yields an exact residual-dependent contribution to $\dot K$ for any differentiable model (Appendix~\ref{app:nonhomogeneous}). We therefore tested whether the homogeneous predictions also hold in a non-homogeneous case, and trained a (42$\times$45)-grid of Transformers. In line with the MLP phase diagram, Figure~\ref{fig:transformer_phase_and_clock}a shows a weak-decay memorization region, an intermediate grokking band, and failure at stronger decay or large learning rate. The fitted curve of the form $\eta=A/\lambda_W$ describes the grokking-memorization boundary well, in line with the homogeneous theory. Additionally, within the grokking networks, Figure~\ref{fig:transformer_phase_and_clock}b exhibits the same linear scaling of the first held-out $80\%$ crossing $s_{80}$ with $(\eta\lambda_W)^{-1}$. We use $80\%$ here because early $10\%$ accuracy crossings are often reversible in practice. This suggests that the optimizer-level phase and transition-time scaling may extend beyond exact homogeneity.

\section{Limitations and Outlook}\label{sec:discussion}
Our theory holds for homogeneous networks trained with squared loss, coupled weight decay, and full-batch gradient flow. Extending the reduced dynamics to more general objectives and architectures is a natural next step. The Transformer experiments already show that similar macroscopic optimizer-space structure can persist without exact homogeneity, but the corresponding microscopic dynamics remain open. Further, the scalar reduction follows a single decoupled task direction. Although the natural Fourier basis of modular addition motivates this, interactions among task-relevant components are neglected, and more general datasets need not provide an equally natural spectral basis. Future work should therefore extend the dynamics to interacting task directions and identify suitable coordinates beyond cyclic modular arithmetic. In this direction, \citet{marchetti2026sequential} show that for sequential group-composition tasks, networks learn irreducible representations one at a time, suggesting a route from the present one-mode reduction toward multiple representation-theoretic components. Finally, modular addition and grokking provide explicit task structure and separate fitting from generalization, but parts of our mechanism are more general. The fixed-kernel residual result does not rely on modular arithmetic, and the residual-dependent contribution to NTK evolution persists for differentiable models. This raises the question of whether our framework also applies to broader representation-learning problems. The J-space framework of~\citet{gurnee2026workspace} provides a possible connection, using an activation-to-output Jacobian to identify representations with strong downstream influence. The NTK similarly uses the parameter Jacobian to track task-aligned directions during training. Establishing whether these directions emerge together would connect the optimization dynamics studied here to the broader formation of internal representations used for generalization.

\section{Conclusions}
We developed a spectral theory of grokking in which coupled weight decay leaves a finite post-fit residual, with target components at smaller NTK eigenvalues retaining a larger residual fraction. We used the neural tangent hierarchy to show that this residual drives task-aligned NTK growth. The resulting slow dynamics determine whether task-aligned NTK strength can grow sufficiently far and sufficiently quickly within the training budget, while training fit and discrete-time stability impose additional constraints. Empirically, a homogeneous MLP on modular addition shows continued Fourier organization of the NTK after training accuracy saturates, an MLP hyperparameter sweep recovers the predicted $(\eta,\lambda_W)$ phase geometry and $(\eta\lambda_W)^{-1}$ transition-time scaling, and a one-block Transformer reproduces the same macroscopic phase structure and transition-time scaling despite lacking exact homogeneity. Together, these results suggest that residual error left after fitting can actively shape how and when neural networks acquire generalizing representations, connecting feature learning, generalization time, and optimizer-space structure within a single mechanistic framework.

\clearpage
\section*{Reproducibility statement}
We provide the proofs and derivations of the homogeneous results in Appendices~\ref{app:fast-proof}--\ref{app:stability} and the differentiable-model identities in Appendix~\ref{app:nonhomogeneous-identities}. Appendix~\ref{app:alignment-experiment} records the complete paired intervention, including architectures, seeds, splits, losses, optimizers, kernel normalization, and reporting convention. Appendices~\ref{app:protocol-spectral}--\ref{app:protocol-potential} give the corresponding details for the spectral diagnostic, MLP and Transformer phase sweeps, boundary calibration, and reduced-mode illustration. The accompanying code\footnote{The code will be made publicly available upon publication.} centralizes the configurations and provides training and analysis entry points.

\section*{AI use statement}
AI was used to assist with text editing, coding, and with literature research. AI was not involved in research ideation, the development of theoretical results, research methodology, and experimental design. All AI-assisted outputs were reviewed, revised where necessary, and verified by the authors.

\clearpage
\bibliographystyle{iclr2027_conference}
\bibliography{references}

\appendix
\section{Related mechanisms for delayed generalization}\label{app:related}
Grokking has been explained through circuit formation, implicit bias, tangent-feature motion, fixed spectra, and optimizer state. Surveys emphasize that several mechanisms can coexist across losses, architectures, and training regimes \citep{bertolotti2026survey}. We position our account by asking which dynamical object changes during the plateau and which observations distinguish that change from a slow fixed-feature relaxation.

\subsection{Mechanistic progress and representation competition}
Modular arithmetic provides an interpretable Fourier basis for circuit analysis. Nanda et al.~\citep{nanda2023progress} reverse-engineer a trigonometric algorithm in a grokked Transformer and introduce progress measures that reveal gradual circuit formation during the apparent plateau. Varma et al.~\citep{varma2024explaining} explain delayed generalization through competition between memorizing and generalizing circuits, while Merrill et al.~\citep{merrill2023tale} study competition between sparse and dense subnetworks. Ding et al.~\citep{ding2024survival} track competing circular Fourier representations with low-dimensional dynamics. He et al.~\citep{he2026modular} analyze how frequency competition, phase alignment, and weight decay shape Fourier-feature formation in two-layer networks on modular addition. Across a broader set of algorithmic Transformer tasks, Naidu et al.~\citep{naidu2026quiet} find causally relevant features developing during long plateaus before the output loss improves. Other representation-level studies connect grokking to changes in norms, learned features, and compression \citep{liu2022towards,liu2023omnigrok,liu2023compression}. We use the same task-defined Fourier basis but study the residual that drives tangent-feature motion within it.

\subsection{Kernel-to-feature transitions and empirical NTK dynamics}
Several theories describe grokking as a transition from an early linearized regime to later feature learning. Kumar et al.~\citep{kumar2024grokking} use a controllable laziness parameter to induce or remove grokking in polynomial regression, MNIST, and modular addition. Lyu et al.~\citep{lyu2024dichotomy} prove a separation between early kernel-like and late implicit biases in homogeneous networks with weight decay. Mohamadi et al.~\citep{mohamadi2024why} show that an early permutation-equivariant kernel regime can be sample inefficient for modular addition, whereas bounded-norm feature-learning solutions generalize from fewer examples. Rubin et al.~\citep{rubin2024phase} analyze an adaptive kernel and relate grokking in two-layer networks to a first-order phase transition. Tian~\citep{tian2026scaling} derives a sequence of lazy, independent-feature, and interacting-feature stages with scaling laws for feature emergence and generalization.

Empirical NTKs distinguish departure from the lazy regime from the task content acquired by the kernel. On sparse parity, Sanguino Bautiste et al.~\citep{sanguino2024feature} observe leading NTK eigenfunctions moving from non-predictive directions toward predictive features as generalization emerges. On image classification, Zheng et al.~\citep{zheng2024delays} find substantial empirical-NTK movement before delayed test improvement and closer synchronization between representational geometry and generalization. For modular arithmetic, Lin~\citep{lin2025feature} shows that leading empirical-NTK eigenspaces align with Fourier feature families in trained MLPs and Transformers and that the alignment changes through grokking. These results establish feature motion but do not identify its source. We derive a residual-dependent equation for task-aligned kernel components. Equation~\eqref{eq:nth} inserts the post-fit residual into $\dot K$, and the Fourier projection yields the reduced feedback, slow time, and finite-training boundary. This mechanism is also compatible with dynamic alignment between finite-network tangent features and task directions \citep{baratin2021alignment}.

\subsection{Static-feature grokking and the weight-decay transition-time scaling}
Fixed-feature models show that delayed generalization can occur without representation change. Levi et al.~\citep{levi2024grokking} analyze grokking in linear estimators, Miller et al.~\citep{miller2024grokking} document related non-neural behavior, and Beck et al.~\citep{beck2025edge} study long delays near linear separability. Xu et al.~\citep{xu2026grok} prove end-to-end grokking for overparameterized ridge regression trained with gradient descent and weight decay. Their bounds separate early training fit from later population improvement and express the delay in terms of learning rate, decay, sample size, feature dimension, and initialization. Kim~\citep{kim2026clock} derives a complementary late-time relaxation in linear models with weight decay and heavy-ball optimization; the zero-momentum limit again yields $1/(\eta\lambda)$ scaling.

The transition-time scaling in Eq.~\eqref{eq:clock} is therefore shared by static-feature and feature-learning mechanisms. Our finite-network account adds the empirical NTK as a state variable and predicts task-aligned motion after training accuracy saturates. The normalized kernel measurement, the residual-dependent hierarchy term, and the directional intervention separate this claim from a fixed-feature delay.

Feature learning can also arise outside ordinary neural-network gradient descent. Mallinar et al.~\citep{mallinar2025emergence} demonstrate grokking in recursive feature machines driven by an average gradient outer product; block-circulant task features emerge after training loss has vanished. Jeffares et al.~\citep{jeffares2024telescoping} analyze grokking through a sequence of local linear approximations, providing another time-resolved view of representation change.

\subsection{Jacobian representations and emergent capabilities}
Jacobian geometry can isolate directions with direct behavioral leverage. Gurnee et al.~\citep{gurnee2026workspace} introduce the Jacobian lens and identify a small ``J-space'' of verbalizable representations in language models; interventions show that these directions can mediate downstream reasoning and reporting. Their Jacobian transports downstream activations to outputs, whereas our NTK uses the parameter Jacobian, but both approaches focus on directions that couple strongly to behavior. We study how such task-relevant parameter-Jacobian directions form during training.

The quantization model of neural scaling \citep{michaud2023quantization} proposes that knowledge and skills may be acquired in discrete modules even when aggregate loss follows smooth power laws. Grokking offers a small setting in which abrupt behavior can be resolved in time. Equation~\eqref{eq:accuracy-cdf} gives one route from a smooth internal coordinate to a sharp output transition, so behavioral discontinuity need not imply discontinuous internal learning. At the same time, task-aligned modes provide a concrete object for testing more modular accounts of capability acquisition.

\subsection{Regularization and optimization mechanisms}
Regularization mechanisms extend beyond Euclidean weight decay. Junior et al.~\citep{junior2025euclidean} show that sparsity, low-rank penalties, implicit regularization, and depth can induce grokking-like transitions. Adaptive optimizers can produce late-training instabilities \citep{thilak2022slingshot}, while cross-entropy training can involve numerical and logit-scaling effects \citep{prieto2025grokking}. The relevant slow variable changes with the objective and optimizer. We focus on homogeneous squared-loss training because coupled $L_2$ decay leaves the ridge residual in Theorem~\ref{thm:fast} and directly contracts the two-homogeneous empirical NTK in Eq.~\eqref{eq:nth}.

\subsection{Relation to fixed-feature and circuit-level accounts}
Our account links the residual left by an early approximately fixed kernel to later tangent-feature motion in a task-defined Fourier direction. The resulting feature strength determines whether held-out margins are reachable and how much optimizer time is required to reach them. Fixed-feature analyses explain how weight decay can create a long transition-time without changing the representation; circuit analyses describe the algorithm implemented after generalization. Residual-dependent neural-tangent-hierarchy dynamics connect the two by specifying how post-fit supervision can reshape the tangent geometry from which a generalizing circuit is built.

\section{Exact homogeneous function-space dynamics}\label{app:fast}
For reference, the notation used throughout the derivations is summarized below.
\begin{table}[H]
\centering
\small
\begin{tabular}{@{}ll@{}}
\toprule
Symbol & Meaning \\
\midrule
$\sig=\ff-\yy$ & stacked training residual \\
$J=\nabla_{\btheta}\ff$ & training-set parameter Jacobian \\
$K=JJ^\top$ & empirical neural tangent kernel \\
$D$ & degree of parameter homogeneity \\
$\Nloss$ & global normalization multiplying $\norm{\sig}^2/2$ in the implemented loss \\
$\lam$ & optimizer coupled $L_2$ weight-decay coefficient \\
$\lamn=\lam/\Nloss$ & normalized decay used in continuous time \\
$\eta$ & optimizer gradient-descent learning rate \\
$t$ & normalized gradient-flow time, $t=\Nloss t_{\rm opt}$ \\
$s$ & discrete optimizer update index \\
$\Lambda$ & task-direction NTK strength $\uu^\top K\uu$ \\
$a$ & projected residual-to-NTK coupling coefficient \\
$y$ & target coefficient in the selected task direction \\
$\kg$ & reduced strength representing a held-out criterion \\
$T$ & finite optimizer update budget \\
\bottomrule
\end{tabular}
\caption{Notation used in the main derivations.}
\label{tab:appendix-notation}
\end{table}

\subsection{Proof of Theorem~\ref{thm:fast}}\label{app:fast-proof}
We absorb the global data-loss prefactor into the normalized time introduced in Section~\ref{sec:fast}. The data-gradient term then has unit coefficient, and optimizer weight decay appears through $\lamn$,
\begin{align}
\dot\btheta
&=-J^\top\sig-\lamn\btheta,\\
\dot\sig
&=J\dot\btheta
=-JJ^\top\sig-\lamn J\btheta
=-K\sig-\lamn J\btheta.
\end{align}
For a $D$-homogeneous network, uniform rescaling of all trainable weights rescales the output by degree $D$. Differentiating this identity at unit scale gives the Euler relation that closes the parameter-space decay term in function space.
\begin{align}
\ff_{c\btheta}&=c^D \ff_{\btheta},\\
\left.\frac{d}{dc}\ff_{c\btheta}\right|_{c=1}
&=J\btheta=D\ff=D(\sig+\yy).
\end{align}
Substituting this relation removes the remaining explicit dependence on $\btheta$.
\begin{equation}
\dot\sig=-(K+D\lamn I)\sig-D\lamn\yy,
\end{equation}
which is Eq.~\eqref{eq:hom-flow}.

For fixed $K$, the residual equation is affine and linear. The matrix $M:=K+D\lamn I$ is positive definite for $\lamn>0$, even when $K$ has null directions, so every residual component relaxes exponentially to a unique fixed point. We set
\[
M:=K+D\lamn I\succ0,\qquad
\sig_*:=-D\lamn M^{-1}\yy,
\]
and the solution can be written as
\begin{align}
\dot\sig&=-M(\sig-\sig_*),\\
\sig(t)-\sig_*&=e^{-Mt}\bigl(\sig(0)-\sig_*\bigr).
\end{align}
We diagonalize $K$ to expose the ridge form. Each eigenmode follows an independent scalar relaxation, and the ratio of its equilibrium residual to its target coefficient is determined by $\Lambda_j$ relative to $D\lamn$. For $K\uu_j=\Lambda_j \uu_j$,
\begin{align}
\dot\sigma_j
&=-(\Lambda_j+D\lamn)\sigma_j-D\lamn y_j,\\
\sigma_{j,*}
&=-\frac{D\lamn}{\Lambda_j+D\lamn}\,y_j.
\end{align}
For ReLU networks, these identities hold on every interval on which the activation pattern is fixed. At activation-boundary crossings the network remains continuous and the gradient-flow equation holds almost everywhere, so the same function-space relation extends piecewise across the trajectory.

This proves Theorem~\ref{thm:fast}. \hfill$\square$

\section{One-mode neural-tangent-hierarchy closure}\label{app:mode}
\subsection{Assumptions behind the modal closure}\label{app:mode-closure}
For the normalized Fourier direction $\uu$ used in the main text, define
\begin{equation}
q_{\btheta}:=\uu^\top\ff_{\btheta},\qquad
\Lambda:=\uu^\top K\uu=\norm{\nabla_{\btheta} q_{\btheta}}^2,\qquad
\sigma:=\uu^\top\sig.
\end{equation}
The reduction follows one task-bearing direction rather than approximating the full kernel. Over the transition interval, we assume that the kernel acts approximately diagonally on this direction and that the residual driving it is dominated by the component along $u$:
\begin{equation}
\sig\approx\sigma \uu,\qquad
K\uu\approx\Lambda \uu.
\label{eq:mode-closure}
\end{equation}
These conditions suppress leading-order mixing with the remaining modes.

With
\begin{equation}
K^{(2)}_{ijk}
=
\left\langle\nabla_{\btheta} K_{ij},\nabla_{\btheta} f_k\right\rangle,
\end{equation}
the projected hierarchy term becomes
\begin{align}
\uu^\top(K^{(2)}\sig)\uu
&\approx
\sigma\sum_{ijk}u_i u_j u_k K^{(2)}_{ijk},\\
a(\btheta)
&:=
\sum_{ijk}u_i u_j u_k K^{(2)}_{ijk}
=
\left\langle\nabla_{\btheta}\Lambda,\nabla_{\btheta} q_{\btheta}\right\rangle.
\label{eq:a-definition}
\end{align}
Thus $a(\btheta)$ is the local proportionality between the residual coefficient $\sigma$ and the residual-driven change in the NTK strength $\Lambda$. When $a>0$ and $\sigma<0$ for a positive target coefficient, this term increases $\Lambda$. The sign of $a$ is a dynamical property; homogeneity does not determine it. The projected kernel equation is therefore
\begin{equation}
\dot\Lambda=-a(\btheta)\sigma-2\lamn\Lambda.
\end{equation}
Whenever $\sigma\neq0$, the same equation can be inverted along a measured trajectory:
\begin{equation}
a_{\rm eff}(t)
=
-\frac{\dot\Lambda(t)+2\lamn\Lambda(t)}{\sigma(t)}.
\label{eq:a-effective}
\end{equation}

The constant-$a$ approximation is local and does not require $a$ to remain fixed throughout training. In the two-homogeneous MLP, both $\Lambda$ and $a$ are degree two under uniform parameter rescaling,
\begin{equation}
\Lambda(c\btheta)=c^2\Lambda(\btheta),
\qquad
a(c\btheta)=c^2a(\btheta).
\end{equation}
Under the one-mode parameter approximation,
\begin{align}
\dot\btheta
&\approx-\sigma\nabla_{\btheta} q_{\btheta}-\lamn\btheta,\\
\dot a
&=-b(\btheta)\sigma-2\lamn a,\qquad
b(\btheta):=
\left\langle\nabla_{\btheta} a,\nabla_{\btheta} q_{\btheta}\right\rangle.
\label{eq:a-dynamics}
\end{align}
After the residual relaxes near its instantaneous ridge value, substitution into Eq.~\eqref{eq:a-dynamics} gives
\begin{equation}
\dot a
\simeq
2\lamn
\left(
\frac{b(\btheta)y}{\Lambda+2\lamn}-a
\right),
\label{eq:a-slow}
\end{equation}
and
\begin{equation}
\frac{d}{dt}\left(\frac{a}{\Lambda}\right)
=
\frac{\sigma}{\Lambda^2}\left(a^2-b\Lambda\right).
\end{equation}
Equations~\eqref{eq:a-dynamics}--\eqref{eq:a-slow} show that weight decay does not introduce an additional fast timescale for $a$ after residual relaxation. Pure radial decay also leaves $a/\Lambda$ unchanged, so any remaining drift of this ratio is residual driven. We therefore use a locally constant coupling subject to four explicit conditions:
\begin{enumerate}[leftmargin=*,itemsep=1pt,topsep=2pt]
\item the projected residual and kernel satisfy Eq.~\eqref{eq:mode-closure};
\item $a(t)$ remains positive over the interval of interest;
\item $a(t)$ changes slowly relative to residual relaxation; and
\item coupling to other task modes is weak enough to be absorbed into the local coefficient $a$ and the effective threshold $\kg$.
\end{enumerate}
The cyclic symmetry of modular addition and the difference-structured split motivate approximate Fourier decoupling. The remaining conditions are assumptions of the scalar reduction.

\subsection{Proof of Theorem~\ref{thm:mode}}\label{app:mode-proof}
With $a$ fixed locally, we combine the two first-order equations into a second-order equation for $\Lambda$. This form separates damping from the force that selects the equilibrium. Equation~\eqref{eq:mode} gives
\begin{align}
\ddot\Lambda
&=-a\dot\sigma-2\lamn\dot\Lambda\\
&=a(\Lambda+2\lamn)\sigma+2a\lamn y-2\lamn\dot\Lambda,\\
\sigma
&=-\frac{\dot\Lambda+2\lamn\Lambda}{a},
\end{align}
hence
\begin{equation}
\ddot\Lambda
+(\Lambda+4\lamn)\dot\Lambda
+2\lamn\Lambda(\Lambda+2\lamn)
-2a\lamn y
=0.
\label{eq:second}
\end{equation}
Equation~\eqref{eq:second} describes one-dimensional damped motion with state-dependent coefficient $\Lambda+4\lamn$. We collect the remaining terms into the cubic potential
\begin{align}
V(\Lambda)
&=
\frac{2\lamn}{3}\Lambda^3
+2\lamn^2\Lambda^2
-2a\lamn y\Lambda,
\label{eq:potential}\\
\mathcal H
&=
\frac12\dot\Lambda^2+V(\Lambda)-V(\Lambda_*).
\label{eq:lyapunov}
\end{align}
On the branch $\Lambda\ge0$, positive decay makes the potential strictly convex. The shifted energy $\mathcal H$ combines displacement from its unique minimum with the kinetic term, and
\begin{align}
V'(\Lambda)
&=2\lamn\Lambda(\Lambda+2\lamn)-2a\lamn y,\\
V''(\Lambda)
&=4\lamn(\Lambda+\lamn)>0,\\
\dot{\mathcal H}
&=-(\Lambda+4\lamn)\dot\Lambda^2\le0.
\end{align}
The stationary points of the first-order system coincide with the extrema of this potential. We solve the equilibrium conditions to obtain
\begin{align}
\sigma_*&=-\frac{2\lamn}{a}\Lambda_*,\\
\Lambda_*(\Lambda_*+2\lamn)&=ay,\\
\Lambda_\pm&=-\lamn\pm\sqrt{\lamn^2+ay}.
\end{align}
For $y=0$, the only nonnegative root is zero. For $y>0$, $\Lambda_+>0$ and $\Lambda_-<0$, giving one positive target-bearing equilibrium. The Lyapunov calculation establishes dissipative motion toward the potential minimum; we use the Jacobian to verify the local asymptotic stability in Theorem~\ref{thm:mode}. At the positive equilibrium,
\begin{equation}
A=
\begin{pmatrix}
-(\Lambda_*+2\lamn)&-\sigma_*\\
-a&-2\lamn
\end{pmatrix},
\end{equation}
with
\begin{align}
\operatorname{tr}A&=-(\Lambda_*+4\lamn)<0,\\
\det A
&=2\lamn(\Lambda_*+2\lamn)-a\sigma_*\\
&=4\lamn(\Lambda_*+\lamn)>0.
\end{align}
The negative trace and positive determinant place both eigenvalues in the open left half-plane, so the positive equilibrium is locally asymptotically stable on the nonnegative branch. \hfill$\square$

\begin{figure}[t]
\centering
\includegraphics[width=0.96\linewidth]{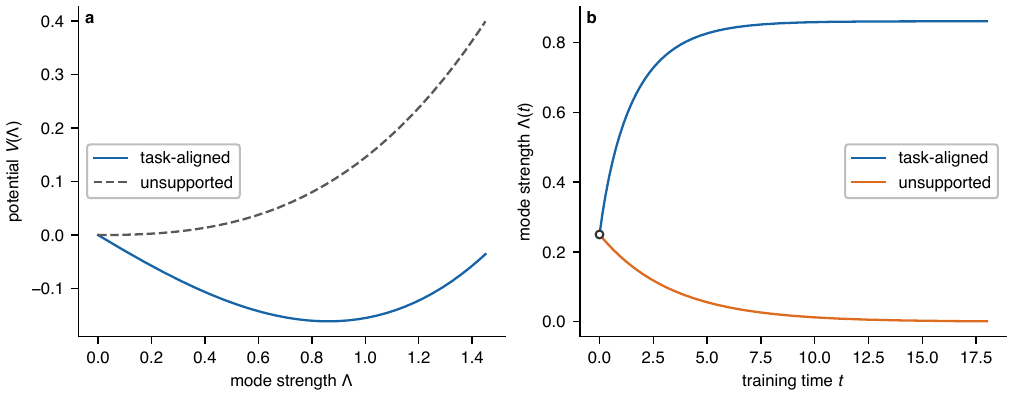}
\caption{Reduced potential and adiabatic trajectories. \textbf{(a)} With illustrative values $\lambda_W^{\mathrm{norm}}=0.15$ and $a=1$, the task-aligned mode ($y=1$) has its potential minimum at positive $\Lambda$, whereas the unsupported mode ($y=0$) is minimized at $\Lambda=0$. \textbf{(b)} Starting from the same initial strength, the two adiabatic trajectories move toward their respective minima. These dimensionless values illustrate the reduced dynamics and are not fitted to an empirical run.}
\label{fig:mode_potential_and_trajectories}
\end{figure}

\section{Adiabatic reduction and finite-training boundary}\label{app:boundary}
\subsection{From smooth mode growth to an accuracy transition}\label{app:boundary-margins}
The reduced dynamics produce a smooth trajectory $\Lambda(t)$, whereas held-out accuracy changes only when individual predictions change class. For held-out example $\mu$ with correct class $c_\mu$, define its correct-class margin along the training trajectory by
\begin{equation}
m_\mu(t)
=
f_{c_\mu}(\xx_\mu;t)
-
\max_{c\neq c_\mu}f_c(\xx_\mu;t).
\end{equation}
When $\Lambda(t)$ is monotone over the transition, the trajectory can be reparameterized by $\Lambda$, so that $\widetilde m_\mu(\Lambda):=m_\mu(t(\Lambda))$. If each relevant $\widetilde m_\mu(\Lambda)$ changes monotonically through its classification transition, then each example has a mode strength at which its predicted class changes. Denote this threshold by
\begin{equation}
\widetilde m_\mu(\Lambda_{\mu,\mathrm g})=0.
\end{equation}
In this case, held-out accuracy at a given $\Lambda$ is the empirical cumulative distribution of the example-specific thresholds,
\begin{equation}
\operatorname{Acc}(\Lambda)
=
\frac1M\sum_{\mu=1}^M
\mathbf 1\{\Lambda>\Lambda_{\mu,\mathrm g}\}
=
\widehat F_{\mathrm g}(\Lambda).
\label{eq:accuracy-cdf}
\end{equation}
A narrow distribution of $\Lambda_{\mu,\mathrm g}$ therefore produces a sharp accuracy rise even when $\Lambda(t)$ changes smoothly. No discontinuity in the optimization dynamics is required. As an illustration, a logistic threshold distribution with location $\mu_g$ and scale $b_g$ gives
\begin{equation}
\operatorname{Acc}(\Lambda)
\approx
\frac{1}{1+\exp[-(\Lambda-\mu_g)/b_g]}.
\end{equation}
This logistic form is only an example of the readout, the one-mode dynamics do not depend on it.

\subsection{Adiabatic slow equation}\label{app:boundary-slow}
The residual relaxes at rate $\Lambda+2\lamn$, while the NTK mode moves on the slower decay-controlled scale once the residual is close to its ridge value. Under this separation of timescales, we set $\dot\sigma\simeq0$ at the current $\Lambda$ and define the resulting adiabatic residual $\sigma_{\rm ad}(\Lambda)$. Equation~\eqref{eq:mode} gives
\begin{align}
0&=-(\Lambda+2\lamn)\sigma-2\lamn y,\\
\sigma_{\rm ad}(\Lambda)&=-\frac{2\lamn y}{\Lambda+2\lamn},\\
\dot\Lambda
&=-a\sigma_{\rm ad}(\Lambda)-2\lamn\Lambda\\
&=
2\lamn
\frac{ay-\Lambda^2-2\lamn\Lambda}{\Lambda+2\lamn}.
\label{eq:slow-app}
\end{align}
The numerator in Eq.~\eqref{eq:slow-app} determines the sign of the mode growth. It is positive below the stable fixed point and vanishes at that fixed point, so the upward motion slows as $\Lambda$ approaches equilibrium. Separating variables gives
\begin{equation}
dt
=
\frac{1}{2\lamn}
\frac{\Lambda+2\lamn}
{ay-\Lambda^2-2\lamn\Lambda}
\,d\Lambda.
\end{equation}
Integrating from $\Lambda_0$ to $\kg$ gives the normalized gradient-flow time required to reach the held-out threshold. After $T$ optimizer updates,
\begin{align}
t&\simeq\Nloss\eta T,\\
\Nloss\lamn&=\lam.
\end{align}
Equating the available normalized time to the crossing time gives Eq.~\eqref{eq:boundary}. The conversion from updates contributes $\Nloss$, while $\Nloss\lamn=\lam$, leaving optimizer decay in the prefactor and normalized decay inside the integrand.

\subsection{Closed form, critical slowing, and the fitting cutoff}\label{app:boundary-closed}
We factor the denominator of the crossing-time integrand at the two fixed points of the slow equation. Define
\begin{equation}
q:=ay,\qquad
\Lambda_\pm:=-\lamn\pm\sqrt{\lamn^2+q}.
\end{equation}
Then
\begin{align}
q-\Lambda^2-2\lamn\Lambda
&=-(\Lambda-\Lambda_+)(\Lambda-\Lambda_-),\\
C_-&:=\frac{2\lamn+\Lambda_+}{\Lambda_+-\Lambda_-},\\
C_+&:=\frac{2\lamn+\Lambda_-}{\Lambda_--\Lambda_+}.
\end{align}
The integrand becomes
\begin{equation}
\frac{\Lambda+2\lamn}{q-\Lambda^2-2\lamn\Lambda}
=
-\frac{C_-}{\Lambda-\Lambda_+}
-\frac{C_+}{\Lambda-\Lambda_-}.
\end{equation}
Integrating between $\Lambda_0$ and $\kg$ gives
\begin{align}
\eta_*(\lam)
=
\frac{1}{2\lam T}\Bigg[
&C_-\log\left|
\frac{\Lambda_0-\Lambda_+}{\kg-\Lambda_+}
\right|
+
C_+\log\left|
\frac{\Lambda_0-\Lambda_-}{\kg-\Lambda_-}
\right|
\Bigg].
\label{eq:closed}
\end{align}
As the requested threshold approaches the stable fixed point $\Lambda_+$, the pole at $\Lambda_+$ reaches the upper integration limit while the $\Lambda_-$ contribution remains finite. Hence, for $\kg\uparrow\Lambda_+$,
\begin{align}
\eta_*(\lam)
&\sim
\frac{C_-}{2\lam T}
\log\frac{1}{\Lambda_+-\kg},\\
\Lambda_+-\kg
&=
\Theta\!\left(\lam^{\mathrm{mode}}-\lam\right)\,.
\end{align}
The logarithm arises because the growth rate of $\Lambda$ vanishes at the reachability boundary. Away from this cutoff, the logarithm remains finite and the explicit prefactor gives the leading $1/(\lam T)$ dependence.

The training-fit cutoff in Eq.~\eqref{eq:fitcut} follows from the same adiabatic gain. For a representative training mode, the expressed fraction of its target component is
\begin{equation}
g(\Lambda_{\rm fit})
=
\frac{\Lambda_{\rm fit}}
{\Lambda_{\rm fit}+2\lamn}.
\end{equation}
Requiring $g(\Lambda_{\rm fit})\ge g_{\min}$ gives
\begin{align}
\frac{\Lambda_{\rm fit}}
{\Lambda_{\rm fit}+2\lamn}
&\ge g_{\min},\\
\lamn
&\le
\frac{\Lambda_{\rm fit}(1-g_{\min})}{2g_{\min}},\\
\lam
&\le
\Nloss\frac{\Lambda_{\rm fit}(1-g_{\min})}{2g_{\min}},
\end{align}
which is Eq.~\eqref{eq:fitcut}. The approximately vertical boundary used in the phase diagram further assumes that the representative training-mode strength $\Lambda_{\rm fit}$ varies weakly with $\eta$ over the range of interest.

\subsection{Discrete-time stability of the frozen-kernel dynamics}\label{app:stability}
Our continuous-time reduction does not describe the large-step edge of the optimizer plane. We hold the kernel fixed to isolate the Euler stability condition for the fast residual dynamics and set
\[
h:=\Nloss\eta.
\]
For fixed $K$,
\begin{align}
\sig_{s+1}
&=
\sig_s-h\left[(K+2\lamn I)\sig_s+2\lamn\yy\right],\\
\sig_*
&=
-2\lamn(K+2\lamn I)^{-1}\yy,\\
\ddelta_s&:=\sig_s-\sig_*,\\
\ddelta_{s+1}
&=
\left[I-h(K+2\lamn I)\right]\ddelta_s.
\end{align}
After we subtract the fixed point, each kernel eigendirection evolves independently. For $K\uu_j=\Lambda_j \uu_j$,
\begin{equation}
\delta_{j,s+1}
=
\left[1-h(\Lambda_j+2\lamn)\right]\delta_{j,s}.
\end{equation}
Therefore
\begin{align}
\left|1-h(\Lambda_j+2\lamn)\right|&<1,\\
\Nloss\eta
&<
\frac{2}{\Lambda_{\max}(K)+2\lamn},\\
\eta
&<
\frac{2}{\Nloss\Lambda_{\max}(K)+2\lam},
\end{align}
which recovers Eq.~\eqref{eq:stabilitycut}. The most restrictive eigenvalue is $\Lambda_{\max}(K)$. Feature learning changes this spectrum, so we use the bound only as a local stability condition for the frozen-kernel approximation, not as a global guarantee for nonlinear training.

\section{Jacobian-mediated feature dynamics beyond homogeneity}\label{app:nonhomogeneous}
For any differentiable model trained with squared loss and coupled $L_2$ decay, the chain rule separates the residual-dependent contribution to tangent-kernel evolution from the contribution of radial parameter decay. Homogeneity is only needed to convert the latter into fixed multiples of the output and NTK. Without homogeneity, the residual-dependent term remains exact, while the radial contractions become architecture dependent.

\subsection{Exact differentiable-model identities}\label{app:nonhomogeneous-identities}
\begin{proposition}[Exact Jacobian-mediated dynamics for differentiable models]\label{prop:general-jacobian}
Let
\begin{equation}
\widetilde{\mathcal L}(\btheta)
=
\frac12\norm{\ff_{\btheta}-\yy}^2
+
\frac{\lamn}{2}\norm{\btheta}^2,
\end{equation}
with $\sig=\ff-\yy$, $J=\nabla_{\btheta}\ff$, and $K=JJ^\top$. Then
\begin{align}
\dot\sig
&=-K\sig-\lamn J\btheta,
\label{eq:general-residual-nonhom}\\
\dot K_{ij}
&=-\sum_kK^{(2)}_{ijk}\sigma_k-\lamn R_{ij},
\label{eq:general-k-nonhom}
\end{align}
where
\begin{equation}
K^{(2)}_{ijk}
=
\left\langle\nabla_{\btheta} K_{ij},\nabla_{\btheta} f_k\right\rangle,
\qquad
R_{ij}
=
\left\langle\nabla_{\btheta} K_{ij},\btheta\right\rangle.
\end{equation}
Along the same flow, the loss is monotone because
\begin{equation}
\frac{d}{dt}\widetilde{\mathcal L}(\btheta(t))
=
-\norm{J^\top\sig+\lamn\btheta}^2
\le0.
\label{eq:general-loss-monotone}
\end{equation}
\end{proposition}

\paragraph{Proof.}
We differentiate the model output along parameter gradient flow to obtain the residual identity. Applying the same derivative to each entry of $K$ produces one contraction with the data gradient and one with the radial weight-decay direction.
\begin{align}
\dot\btheta
&=-J^\top\sig-\lamn\btheta,\\
\dot\sig
&=J\dot\btheta
=-K\sig-\lamn J\btheta,\\
\dot K_{ij}
&=
\left\langle\nabla_{\btheta} K_{ij},\dot\btheta\right\rangle\\
&=
-\sum_k
\left\langle\nabla_{\btheta} K_{ij},\nabla_{\btheta} f_k\right\rangle\sigma_k
-\lamn
\left\langle\nabla_{\btheta} K_{ij},\btheta\right\rangle,\\
\frac{d}{dt}\widetilde{\mathcal L}
&=
\left\langle
\nabla_{\btheta}\widetilde{\mathcal L},
\dot\btheta
\right\rangle
=
-\norm{\nabla_{\btheta}\widetilde{\mathcal L}}^2.
\end{align}
\hfill$\square$

Equation~\eqref{eq:general-k-nonhom} isolates the extension needed beyond homogeneous networks. Whenever the $K^{(2)}$ contraction is nonzero, the residual that controls prediction error also changes the tangent kernel. The contraction may vanish at individual states; homogeneity is needed only to replace the remaining radial term by fixed coefficients.

\subsection{Local projected dynamics}\label{app:nonhomogeneous-local}
For a task direction $\uu$, we suppose over a training interval that
\begin{equation}
K\uu\approx\Lambda \uu,\quad
\sig\approx\sigma \uu,\quad
\uu^\top J\btheta\approx\beta(\sigma+y),\quad
\uu^\top R \uu\approx\rho\Lambda,\quad
\uu^\top(K^{(2)}\sig)\uu\approx a\sigma,
\label{eq:nonhom-local-assumptions}
\end{equation}
with slowly varying $\beta,\rho,a$. The coefficients $\beta$ and $\rho$ summarize radial parameter decay after projection onto the local task direction, while $a$ retains its residual-to-kernel role. Projecting gives
\begin{equation}
\dot\sigma
\approx
-(\Lambda+\beta\lamn)\sigma-\beta\lamn y,
\qquad
\dot\Lambda
\approx
-a\sigma-\rho\lamn\Lambda.
\label{eq:nonhom-local-mode}
\end{equation}
For a two-homogeneous model, Euler identities set $\beta=\rho=2$ and recover Eq.~\eqref{eq:mode}. In a non-homogeneous model, we must measure these local contractions; they need not equal the homogeneous coefficients. If they remain order one and vary slowly through the transition, both decay terms retain the factor $\lamn$, leaving $1/\lamn$ as the natural normalized-time scale and $(\eta\lambda_W)^{-1}$ as the corresponding update scale.

For the Transformer in Appendix~\ref{app:protocol-transformer}, we let $\ff_{\btheta}$ denote the softmax probabilities entering the mean-squared-error loss. Proposition~\ref{prop:general-jacobian} then applies despite attention, LayerNorm, residual paths, and biases.

\section{Directional NTK-alignment intervention}\label{app:alignment}
\subsection{Earlier task alignment shortens the predicted delay}\label{app:alignment-prediction}
Starting closer to the held-out threshold removes a positive part of the crossing-time integral without invoking the global kinetic calibration. Equation~\eqref{eq:boundary} gives
\begin{equation}
s_{\mathrm g}
=
\frac{1}{2\eta\lam}
\int_{\Lambda_0}^{\kg}
\frac{\Lambda+2\lamn}
{ay-\Lambda^2-2\lamn\Lambda}
\,d\Lambda,
\label{eq:grok-time-initial}
\end{equation}
and differentiation with respect to the initial strength gives
\begin{equation}
\frac{\partial s_{\mathrm g}}{\partial\Lambda_0}
=
-\frac{1}{2\eta\lam}
\frac{\Lambda_0+2\lamn}
{ay-\Lambda_0^2-2\lamn\Lambda_0}
<0
\label{eq:initial-alignment}
\end{equation}
throughout the reachable branch, where the denominator is the positive spectral drive. The sign predicts that improving early alignment with a reachable task direction should advance the later held-out transition even if the auxiliary intervention is removed before generalization begins.

\subsection{A temporary NTK intervention advances held-out generalization}\label{app:alignment-experiment}
We change early task-indexed tangent geometry while keeping the later objective identical, which tests the sign in Eq.~\eqref{eq:initial-alignment}. We use addition modulo $97$ on the $4{,}753$ unordered input pairs. We train a bias-free $194\!\to\!128\!\to\!128\!\to\!128\!\to\!97$ ReLU teacher with He-normal initialization. Its $50/50$ split uses NumPy seed $0$ and contains $2{,}376$ training and $2{,}377$ held-out examples. We run $30{,}000$ full-batch SGD updates without momentum, using learning rate $180$, coupled weight decay $10^{-5}$, ordinary PyTorch MSE between logits and one-hot targets averaged over samples and outputs, and model seed $0$.

For examples $\xx_\mu,\xx_\nu$ with labels $c_\mu,c_\nu$, we project the empirical NTK through the corresponding true-label logits,
\begin{equation}
K^{\rm TL}_{\mu\nu}
=
\left\langle
\nabla_{\btheta} f_{c_\mu}(\xx_\mu),
\nabla_{\btheta} f_{c_\nu}(\xx_\nu)
\right\rangle.
\label{eq:true-label-intervention}
\end{equation}
The intervention uses a normalization order different from the diagnostic in Appendix~\ref{app:protocol-spectral}. We first cosine-normalize the sample kernel,
\begin{equation}
\widehat K^{\rm TL}_{\mu\nu}
=
\frac{K^{\rm TL}_{\mu\nu}}
{\sqrt{K^{\rm TL}_{\mu\mu}K^{\rm TL}_{\nu\nu}}},
\end{equation}
and then average between label groups,
\begin{equation}
\bar K_{cc'}
=
\frac{1}{n_c n_{c'}}
\sum_{\mu:c_\mu=c}\sum_{\nu:c_\nu=c'}
\widehat K^{\rm TL}_{\mu\nu}.
\label{eq:intervention-label-kernel}
\end{equation}
We compute the fixed symmetric $97\times97$ target matrix from the teacher's training split and the differentiable student matrix from each student's training split.

We train $40$ paired bias-free $194\!\to\!256\!\to\!97$ ReLU students with He-normal initialization. Pair $q\in\{0,\ldots,39\}$ uses model seed $q$ and an independently generated $50/50$ split with NumPy seed $10{,}000+q$; every split has $2{,}376$ training and $2{,}377$ held-out examples, and both members of the pair share the same initialization and data. Every student uses $30{,}000$ full-batch SGD updates without momentum, learning rate $35$, coupled weight decay $3\times10^{-6}$, and ordinary PyTorch MSE between logits and one-hot targets averaged over samples and outputs. During updates $0$ through $2{,}999$, the intervention member minimizes
\begin{equation}
\mathcal L_{\rm int}
=
\mathcal L_{\rm student}
+
\lambda_{\rm NTK}
\left\|\bar K_{\btheta}-\bar K_{\rm teacher}\right\|_F,
\qquad
\lambda_{\rm NTK}=0.10.
\label{eq:ntk-align}
\end{equation}
We remove the auxiliary term at update $3{,}000$; the baseline never receives it. We record training and held-out accuracy and MSE every $100$ updates.

\begin{figure}[t]
\centering
\includegraphics[width=\linewidth]{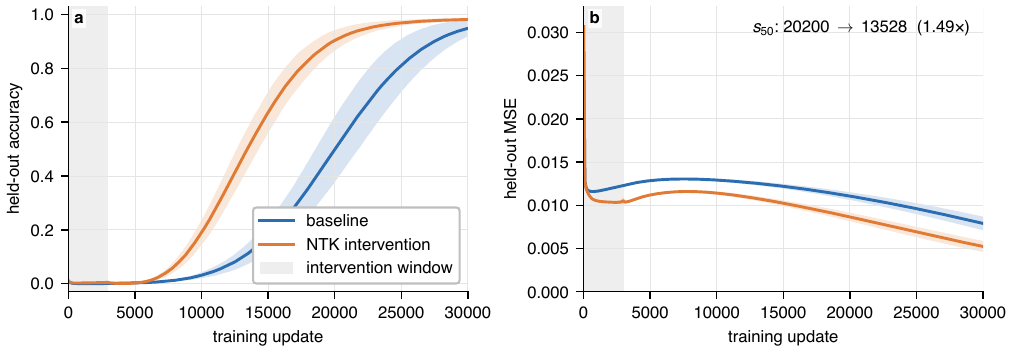}
\caption{Temporary early NTK alignment advances held-out generalization. \textbf{(a)} Mean held-out accuracy for the paired baseline and intervention students. The shaded vertical region marks the first $3{,}000$ updates, when the NTK penalty is active; bands show one sample standard deviation across $40$ paired seeds. \textbf{(b)} Mean held-out MSE for the same runs. The mean first $50\%$ held-out crossing moves from $20{,}200$ updates for the baseline to $13{,}527.5$ for the intervention, a ratio of $1.493$.}
\label{fig:alignment_intervention}
\end{figure}

The paired mean speedup is $1.498$, and the median paired speedup is $1.489$. The intervention students also maintain lower held-out MSE after the penalty has been removed. Because the two members of each pair share their data and initialization and use the same objective after update $3{,}000$, the shift is consistent with the predicted negative derivative in Eq.~\eqref{eq:initial-alignment}. Matching the full $97\times97$ label kernel changes several Fourier components and may alter other parts of the trajectory, so the experiment tests the direction of the initial-alignment effect rather than the scalar one-mode closure in isolation.

\section{Experimental configurations and analysis conventions}\label{app:protocols}
The accompanying experiment package centralizes the configurations below in the model, training, and analysis code. We report optimizer weight decay in the convention passed directly to PyTorch SGD. The anonymous artifact URL remains to be inserted in the reproducibility statement.

\subsection{Spectral diagnostics on addition modulo 97}\label{app:protocol-spectral}
For Figure~\ref{fig:ntk_accuracy_eigenvectors}, we use the $97\cdot98/2=4{,}753$ unordered input pairs allowed by commutativity. A NumPy generator with seed $0$ permutes the pairs; the first $2{,}376$ are used for training and the remaining $2{,}377$ are held out. Each input concatenates two $97$-dimensional one-hot vectors, and the target is the $97$-dimensional one-hot encoding of the modular sum.

The model is a bias-free $194\!\to\!256\!\to\!97$ one-hidden-layer ReLU MLP with He-normal initialization and model seed $0$. We train for $20{,}000$ full-batch SGD updates with learning rate $40$, coupled weight decay $2.5\times10^{-5}$, and no momentum. The loss is ordinary PyTorch MSE averaged over samples and all $97$ output coordinates, so $\Nloss=2/(2376\cdot97)=1/115{,}236$ and $\lamn=2.8809$. Accuracy is recorded every $100$ updates.

For training examples $\xx_\mu,\xx_\nu$ with labels $c_\mu,c_\nu$, the true-label-projected sample NTK is
\begin{equation}
K^{\rm TL}_{\mu\nu}
=
\left\langle
\nabla_{\btheta} f_{c_\mu}(\xx_\mu),
\nabla_{\btheta} f_{c_\nu}(\xx_\nu)
\right\rangle.
\end{equation}
We first average these entries between groups of examples with the same output labels, producing a $97\times97$ matrix $\widetilde K$. We then diagonal-normalize and symmetrize the label kernel,
\begin{equation}
\bar K_{cc'}
=
\frac{\widetilde K_{cc'}}
{\sqrt{\widetilde K_{cc}\widetilde K_{c'c'}}}.
\end{equation}
Panel (a) uses updates $0$, $3{,}000$, $8{,}000$, and $20{,}000$. Eigensystems are computed at updates $0$, $100$, $500$, $1{,}000$, $3{,}000$, $6{,}000$, $8{,}000$, $10{,}000$, and $20{,}000$. Eigenvectors are matched backward through these checkpoints by maximum absolute overlap, with signs chosen continuously. Panel (c) displays four tracked leading eigenvectors at updates $0$, $1{,}000$, $6{,}000$, and $20{,}000$.

\subsection{High-resolution MLP phase sweep on addition modulo 23}\label{app:protocol-mlp}
We use all $23^2=529$ ordered input pairs. Each input concatenates two $23$-dimensional one-hot vectors, and the target is a $23$-dimensional one-hot encoding of the modular sum. NumPy seed $2027$ selects the training differences
\begin{equation}
\mathcal{D}=\{0,1,2,3,4,5,6,7,12,13,14,16,17,19,20,22\},
\end{equation}
and we define
\begin{equation}
\mathcal T_{\mathcal{D}}=\{(a,b):a-b\bmod23\in \mathcal D\}.
\end{equation}
The split contains $368$ training and $161$ held-out pairs and is invariant under simultaneous shifts $(a,b)\mapsto(a+r,b+r)$.

At every grid point we initialize the same bias-free $46\!\to\!256\!\to\!23$ one-hidden-layer ReLU MLP with He-normal initialization and model seed $0$. Training uses ordinary PyTorch MSE averaged over the $368\times23$ residual coordinates, full-batch SGD, coupled $L_2$ weight decay, and no momentum. Thus $\Nloss=2/(368\cdot23)=1/4232$. Each run lasts $24{,}000$ updates and is evaluated every $100$ updates. The grid contains $84$ logarithmically spaced learning rates from $0.1$ to $100$ and $90$ logarithmically spaced weight decays from $5\times10^{-6}$ to $2\times10^{-3}$, for $7{,}560$ runs.

We classify runs using a $90\%$ accuracy threshold. \emph{Grokking} requires final training and held-out accuracy above $90\%$; \emph{memorization} requires final training accuracy above $90\%$ but held-out accuracy below it; \emph{forgetting} means that training accuracy crossed $90\%$ earlier but ends below it; all remaining runs are \emph{no fitting}. For Figure~\ref{fig:mlp_phase_and_clock}b, we retain only runs that ultimately grok and record the first update at which held-out accuracy reaches $10\%$. This lower threshold marks the beginning of the final MLP generalization rise and does not affect the phase labels.

\begin{figure}[t]
\centering
\includegraphics[width=0.96\linewidth]{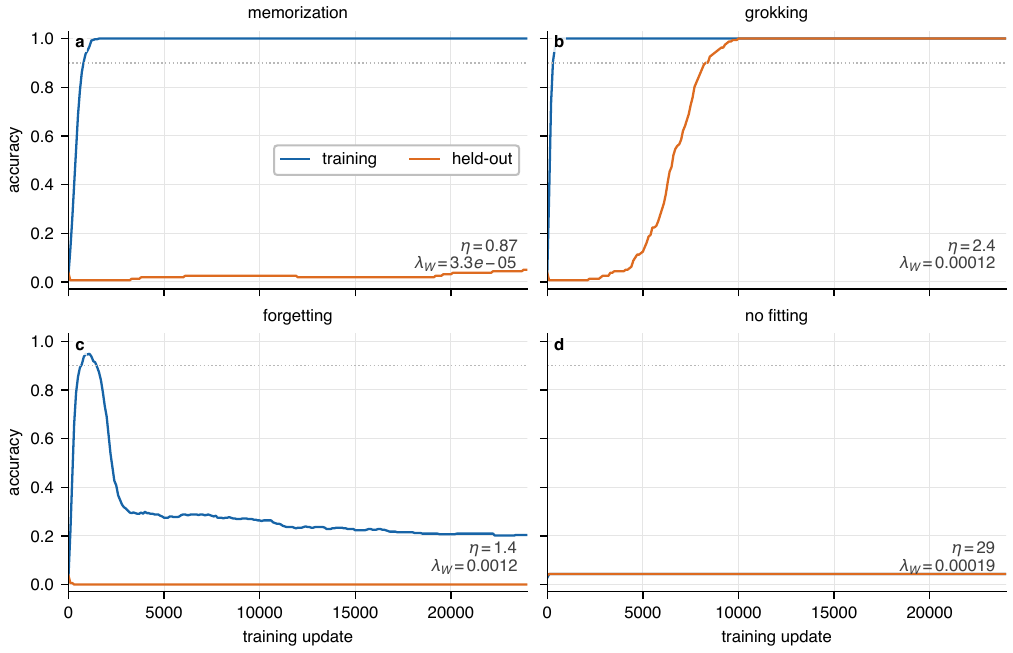}
\caption{Representative trajectories for the four MLP phase outcomes. We show training and held-out accuracy for fixed cells from the modulo-$23$ sweep. The dotted horizontal line is the $90\%$ threshold used to classify the phase map.}
\label{fig:mlp_phase_example}
\end{figure}

The representative runs in Figure~\ref{fig:mlp_phase_example} are fixed cells from the same dense sweep. Their $(\eta,\lam)$ values are $(0.87048,3.2930\times10^{-5})$ for memorization, $(2.36316,1.18329\times10^{-4})$ for grokking, $(1.43426,1.24846\times10^{-3})$ for forgetting, and $(28.6967,1.89561\times10^{-4})$ for no fitting. The grokking example first reaches $90\%$ held-out accuracy at update $8{,}300$. The forgetting example reaches peak training accuracy $0.948$ and ends at $0.204$.

\subsubsection{Boundary extraction and calibration}\label{app:protocol-mlp-calibration}
For each usable weight-decay column, we define the memorization--grokking boundary as the geometric midpoint between the last memorizing learning rate and the first grokking learning rate. We exclude the detached high-step island. The finite-time boundary uses $53$ midpoints over $2\times10^{-5}\le\lam\le7\times10^{-4}$ with equal weight in log learning rate.

To evaluate Eq.~\eqref{eq:boundary}, define the scaled coordinates
\begin{equation}
\widehat\Lambda=\Nloss\Lambda,
\qquad
\widehat q=\Nloss^2ay.
\end{equation}
This removes $\Nloss$ from the integrand while leaving the optimizer prefactor $1/(2\lam T)$ unchanged. The calibrated finite-time curve is
\begin{equation}
\boxed{
\eta_{\rm fit}(\lam)
=
\frac{C_\eta}{2\lam T}
\int_{\widehat\Lambda_0}^{\widehat\Lambda_g}
\frac{\widehat\Lambda+2\lam}
{\widehat q-\widehat\Lambda^2-2\lam\widehat\Lambda}
\,d\widehat\Lambda
}
\label{eq:calibrated-boundary-app}
\end{equation}
with one global kinetic factor $C_\eta$. If omitted modes and slow variation in the projected coupling rescale the scalar velocity as $d\widehat\Lambda/dt\simeq\kappa F(\widehat\Lambda)$, then crossing times are multiplied by $1/\kappa$ and $C_\eta=1/\kappa$. The MSE normalization $\Nloss$ is known exactly and is not part of this calibration. The quantities $\widehat\Lambda_0$, $\widehat\Lambda_g$, and $\widehat q$ are inferred jointly from the optimizer-plane boundary rather than measured from NTK trajectories. We therefore treat them as effective coordinates of the scalar boundary model: the fit tests the predicted dependence of the crossing time on learning rate and weight decay, but does not determine the absolute magnitude of post-fit NTK growth.

Panel (a) of Table~\ref{tab:boundary-calibration} collects the values used for every overlay in Figure~\ref{fig:mlp_phase_and_clock}a. The mode-reachability cutoff is derived from the fitted reduced coordinates rather than fitted independently. The training-fit and stability overlays use the functional forms of Eqs.~\eqref{eq:fitcut} and \eqref{eq:stabilitycut}, with their locations calibrated separately to the observed boundaries. In particular, the stability calibration is summarized by an effective scale $\widehat\Lambda_{\rm stab}$ and is not obtained from a direct measurement of $\Nloss\Lambda_{\max}(K)$.

Over the small-decay interval used for the inverse-decay comparison, the empirical boundary has log--log slope $-1.1681$, while the calibrated finite-time curve has slope $-0.9043$. These slopes are diagnostics of the boundary shape and are not additional fit parameters.

\subsection{Transformer phase sweep on addition modulo 23}\label{app:protocol-transformer}
We use the same task, difference split, split seed, training size, and held-out size as in Appendix~\ref{app:protocol-mlp}, but present each input as a sequence of two $23$-dimensional one-hot tokens. A learned affine encoder maps each token to $d_{\rm model}=32$. The model contains one pre-norm Transformer block with four attention heads, a ReLU feed-forward width of $64$, residual connections, learned embeddings for the two positions, a final LayerNorm, and an affine $64\!\to\!23$ decoder. We use model seed $0$ at every grid point.

Training uses full-batch SGD with coupled $L_2$ weight decay and no momentum. The loss is PyTorch MSE between softmax probabilities and one-hot targets, averaged over samples and output coordinates, so $\Nloss=1/4232$. Each run lasts $60{,}000$ updates and is evaluated every $250$ updates. The phase file contains $42$ uniformly logarithmically spaced learning rates from $5\times10^{-3}$ to $10^3$ and $45$ uniformly logarithmically spaced weight decays from $10^{-7}$ to $10^{-3}$, for $1{,}890$ independently trained networks. We use the same $90\%$ phase definitions as for the MLP.

For every usable weight-decay column, we extract the geometric midpoint between the last memorizing and first grokking learning rate. We fit the inverse-decay form $\eta=A/\lam$ over $10^{-5}\le\lam\le3\times10^{-4}$. The current phase grid yields the values in panel (b) of Table~\ref{tab:boundary-calibration}.

\begin{table}[t]
\centering
\small
\setlength{\tabcolsep}{4pt}
\begin{tabular}{@{}p{2.15cm}p{2.55cm}p{4.65cm}p{2.95cm}@{}}
\toprule
Curve & Relation & Calibration values & Fit range / quality \\
\midrule
\multicolumn{4}{@{}l}{\textbf{(a) MLP}} \\
\addlinespace[2pt]
Finite-time boundary
& Eq.~\eqref{eq:calibrated-boundary-app}
& $C_\eta=24.840743$, $\widehat\Lambda_0=0.9281806$, $\widehat\Lambda_g=0.9284920$, $\widehat q=0.8633973$; $\kappa=C_\eta^{-1}=0.0402564$
& $53$ midpoints, $2\times10^{-5}\le\lam\le7\times10^{-4}$; log-MSE $0.026800$; log-RMSE $0.163706$ \\
\addlinespace
Inverse-decay asymptote
& $\eta=A/\lam$
& $A=1.0916395\times10^{-4}$
& $34$ midpoints, $2\times10^{-5}\le\lam\le2\times10^{-4}$; log-RMSE $0.119902$ \\
\addlinespace
Mode reachability
& Eq.~\eqref{eq:featurecut}
& $\lam^{\rm mode}=7.0000\times10^{-4}$, derived from $\widehat q$ and $\widehat\Lambda_g$
& no independent fit \\
\addlinespace
Training fit
& Eq.~\eqref{eq:fitcut}
& $\lam^{\rm fit}=8.3821151\times10^{-4}$
& location minimizes final-training-accuracy misclassification \\
\addlinespace
Frozen-kernel stability
& Eq.~\eqref{eq:stabilitycut}
& effective $\widehat\Lambda_{\rm stab}=0.1056011$; low-decay cutoff $\eta\simeq18.9282$
& location calibrated to observed instability boundary \\
\midrule
\multicolumn{4}{@{}l}{\textbf{(b) Transformer}} \\
\addlinespace[2pt]
Inverse-decay boundary
& $\eta=A/\lam$
& $A=1.6875621\times10^{-5}$
& $16$ boundary points, $10^{-5}\le\lam\le3\times10^{-4}$; log-RMSE $0.107016$ \\
\bottomrule
\end{tabular}
\caption{Boundary calibrations for the optimizer-plane experiments. Panel (a) lists the MLP phase-diagram overlays; panel (b) gives the inverse-decay fit for the Transformer memorization--grokking boundary. The theory specifies the functional forms of the MLP curves, while their locations are determined either by the reduced finite-time fit or by separate effective calibrations to the observed boundaries.}
\label{tab:boundary-calibration}
\end{table}

For the transition-time scaling panel, we retain runs that ultimately grok and record the first held-out $80\%$ crossing. The $80\%$ threshold is chosen to lie in the final generalization rise rather than near chance accuracy, $1/23\approx4.35\%$. On the current $1{,}890$-run grid, the grokking cohort has median $\eta\lam s_{80}=0.920$.

\subsection{Reduced-mode potential illustration}\label{app:protocol-potential}
For Figure~\ref{fig:mode_potential_and_trajectories}, we integrate Eq.~\eqref{eq:slow} with dimensionless normalized decay $\lamn=0.15$, coupling $a=1$, initial strength $\Lambda(0)=0.25$, fourth-order Runge--Kutta step $0.002$, and final time $18$. The task-aligned trajectory uses $y=1$, while the unsupported control uses $y=0$. These values are chosen only to display the qualitative reduced dynamics and are not fitted to an empirical run.

\end{document}